\documentclass[runningheads]{llncs}
\usepackage[T1]{fontenc}
\usepackage[utf8]{inputenc} % allow utf-8 input
\usepackage[T1]{fontenc}    % use 8-bit T1 fonts
\usepackage{url}            % simple URL typesetting
\usepackage{booktabs}       % professional-quality tables
\usepackage{subcaption}
\usepackage{amsfonts}       % blackboard math symbols
\usepackage{amsmath}

\usepackage{algorithm}
\usepackage{algorithmicx}
\usepackage{algpseudocode}

\usepackage{hyperref}
\usepackage{cleveref}

\usepackage{multirow}
\usepackage{xcolor}
\usepackage{nicefrac}       % compact symbols for 1/2, etc.
\usepackage{microtype}      % microtypography
\usepackage{lipsum}
\usepackage{graphicx}
\usepackage{color}

\DeclareCaptionLabelFormat{algnonumber}{Algorithm}
\newcommand{\eg}{e.g., }

\begin{document}
\title{iPINNs: Incremental learning for Physics-informed neural networks}
\author{Aleksandr Dekhovich\inst{1} \and
Marcel H.F. Sluiter \inst{1} \and
David M.J. Tax \inst{2} \and Miguel A. Bessa \inst{3} }
\authorrunning{A. Dekhovich et al.}
\institute{Department of Materials Science and Engineering, Delft University of Technology, Mekelweg 2, Delft, 2628 CD,  The Netherlands \and
Pattern Recognition and Bioinformatics Laboratory, Delft University of Technology, Van Mourik Broekmanweg 6, Delft, 2628 XE, The Netherlands \and
School of Engineering, Brown University, 184 Hope St., Providence, RI 02912, USA\\
\email{miguel\_bessa@brown.edu} }
%

%\author{Anonymous authors}
%\institute{Affiliation}

\maketitle              % typeset the header of the contribution
\begin{abstract}
Physics-informed neural networks (PINNs) have recently become a powerful tool for solving partial differential equations (PDEs). However, finding a set of neural network parameters that lead to fulfilling a PDE can be challenging and non-unique due to the complexity of the loss landscape that needs to be traversed. Although a variety of multi-task learning and transfer learning approaches have been proposed to overcome these issues, there is no incremental training procedure for PINNs that can effectively mitigate such training challenges. We propose incremental PINNs (iPINNs) that can learn multiple tasks (equations) sequentially without additional parameters for new tasks and improve performance for every equation in the sequence. Our approach learns multiple PDEs starting from the simplest one by creating its own subnetwork for each PDE and allowing each subnetwork to overlap with previously learned subnetworks. We demonstrate that previous subnetworks are a good initialization for a new equation if PDEs share similarities. We also show that iPINNs achieve lower prediction error than regular PINNs for two different scenarios: (1) learning a family of equations (\eg 1-D convection PDE); and (2) learning PDEs resulting from a combination of processes (\eg 1-D reaction-diffusion PDE). The ability to learn all problems with a single network together with learning more complex PDEs with better generalization than regular PINNs will open new avenues in this field.

\end{abstract}

\section{Introduction}
Deep neural networks (DNNs) play a central role in scientific machine learning (SciML). Recent advances in neural networks find applications in real-life problems in physics \cite{madrazo2019application,bogatskiy2020lorentz,shlomi2020graph,khatib2022learning}, medicine \cite{marques2020automated,si2022artificial,sarvamangala2022convolutional}, finance \cite{hosaka2019bankruptcy,yu2020stock,gogas2021machine,wang2023neurodynamics}, and engineering \cite{bessa2017framework,sosnovik2019neural,chandrasekhar2021tounn,juan2022review}. In particular, they are also applied to solve Ordinary Differential Equations and Partial Differential Equations (ODEs/PDEs) \cite{lee1990neural,meade1994solution,yentis1996vlsi,raissi2019physics}.  Consider the following PDE,

\begin{align}\label{eq:pde}
    \mathcal{F}[u(\mathbf{x}, t)] &= f(\mathbf{x}), \quad \mathbf{x} \in \mathrm{\Omega}, \ t \in [t_0, T], \\
    \mathcal{B}[u(\mathbf{x}, t)] &= b(\mathbf{x}), \quad \mathbf{x} \in  \mathrm{\partial\Omega}, \\
    u(\mathbf{x} , t_0) &= h(\mathbf{x}), \quad \mathbf{x} \in \mathrm{\Omega},
\end{align}

\noindent where $\mathcal{F}$ is a differential operator, $\mathcal{B}$ is a boundary condition operator, $h(\mathbf{x})$ is an initial condition, and $\mathrm{\Omega}$ is a bounded domain. %The first works in solving ODEs and PDEs with neural networks were published in the 90s of the 20th century \cite{lee1990neural,meade1994solution,yentis1996vlsi}. These methods were based on the discretization of the domain and the following solving of the linear system of algebraic equations. 

The first neural network-based approaches incorporated a form of the equation into the loss function with initial and boundary conditions included as hard constraints \cite{lagaris1997artificial,lagaris1998artificial}. However, these works used relatively small neural networks with one or two hidden layers. On the contrary, PINNs \cite{raissi2019physics} encode initial and boundary conditions as soft constraints into the loss function of a DNN. Subsequently, PINNs and their extensions found applications in fluid mechanics \cite{mao2020physics,wessels2020neural,cai2022physics}, inverse problems \cite{chen2020physics,lu2021physics,wiecha2021deep} and finance \cite{raissi2019physics,bai2022application}. Later, the generalized version of PINNs, called XPINNs \cite{jagtap2021extended}, was proposed by decomposing the domain into multiple subdomains. However, this method uses as many networks as the number of subdomains, increasing the algorithm's complexity. Multi-head PINNs (MH-PINNs) \cite{zou2023hydra} is a multi-task learning approach for PINNs that is employed to learn stochastic processes, synergistic learning of PDEs and uncertainty quantification. MH-PINNs have a shared part of the network and task-specific output heads for prediction. Therefore, it uses additional parameters for every head, increasing the model's size with respect to the number of tasks.

Despite the popularity of DNNs, and PINNs in particular, there are few \textit{incremental} learning algorithms available in SciML literature. Yet, incremental learning and continual learning algorithms \cite{castro2018end,cermelli2022modeling,kang2022class} are capable of handling tasks sequentially, instead of altogether as in multi-task learning and other strategies. Moreover, they are still capable of not forgetting how to solve all of the previously learned tasks. If tasks have some similarities with each other, new tasks have the potential of being learned better (i.e., faster or with lower testing error) with the help of previously learned ones. The goal of this work is to propose an incremental learning algorithm for PINNs such that similar symbiotic effects can be obtained.

\paragraph{\bf Background and main challenges.} 

PINNs formulate the PDE solution problem by including initial and boundary conditions into the loss function of a neural network as soft constraints. Let us denote the output of the network $\mathcal{N}$ with learnable parameters $\theta$ as $\hat{u}(\theta, \mathbf{x}, t) = \mathcal{N}(\theta; \mathbf{x}, t)$. Then sampling the set of collocation points, i.e.  a set of points in the domain, $\mathcal{CP} = \{ (x^i, t^i) : x^i\in \text{int }\mathrm{\Omega}, \ t^i \in (t_0, T], \ i = 1, 2, \ldots N_{\mathcal{F}} \}$, the set of initial points $\mathcal{IP} = \{(x^j, t_0): x^j \in \partial \mathrm{\Omega}, \ j = 1, 2, \ldots, N_{u_0} \}$ and the set of boundary points $\mathcal{BP} = \{(x^k, t^k): x^k \in \partial \mathrm{\Omega}, \ t^k \in (t_0, T], \ k = 1, 2, \ldots, N_b \}$ one can write the optimization problem and loss function arising from PINNs as follows:

\begin{align}\label{eq:pinns_loss}
    \mathcal{L}(\theta) &= \mathcal{L}_{\mathcal{F}} (\theta) + \mathcal{L}_{u_0} (\theta) + \mathcal{L}_{b} (\theta) \to \min_{\theta},\\
    \mathcal{L}_{\mathcal{F}} (\theta) &= \frac{1}{N_{\mathcal{F}}} \sum_{i=1}^{N_{\mathcal{F}}} \big\lvert \big\lvert \mathcal{F}[\hat{u}(\theta, x^i, t^i)] - f(x^i) \big\rvert \big\rvert^2, \quad (x^i, t^i) \in \mathcal{CP}, \\
    \mathcal{L}_{u_0} (\theta) &= \frac{1}{N_{u_0}} \sum_{j=1}^{N_{u_0}} \big\lvert \big\lvert \hat{u}(\theta, x^j, t_0) - h(x^j) \big\rvert \big\rvert^2, \quad (x^j, t_0) \in \mathcal{IP}, \\
    \mathcal{L}_{b}(\theta) &= \frac{1}{N_b} \sum_{k=1}^{N_b} \big\lvert \big\lvert \mathcal{B}[\hat{u}(\theta, x^k, t^k)] - b(x^k) \big\rvert \big\rvert^2, \quad (x^k, t^k) \in \mathcal{BP}.
\end{align}

\begin{figure}[ht!]
    \begin{subfigure}{0.4\textwidth}
        \centering
        \includegraphics[width=\textwidth]{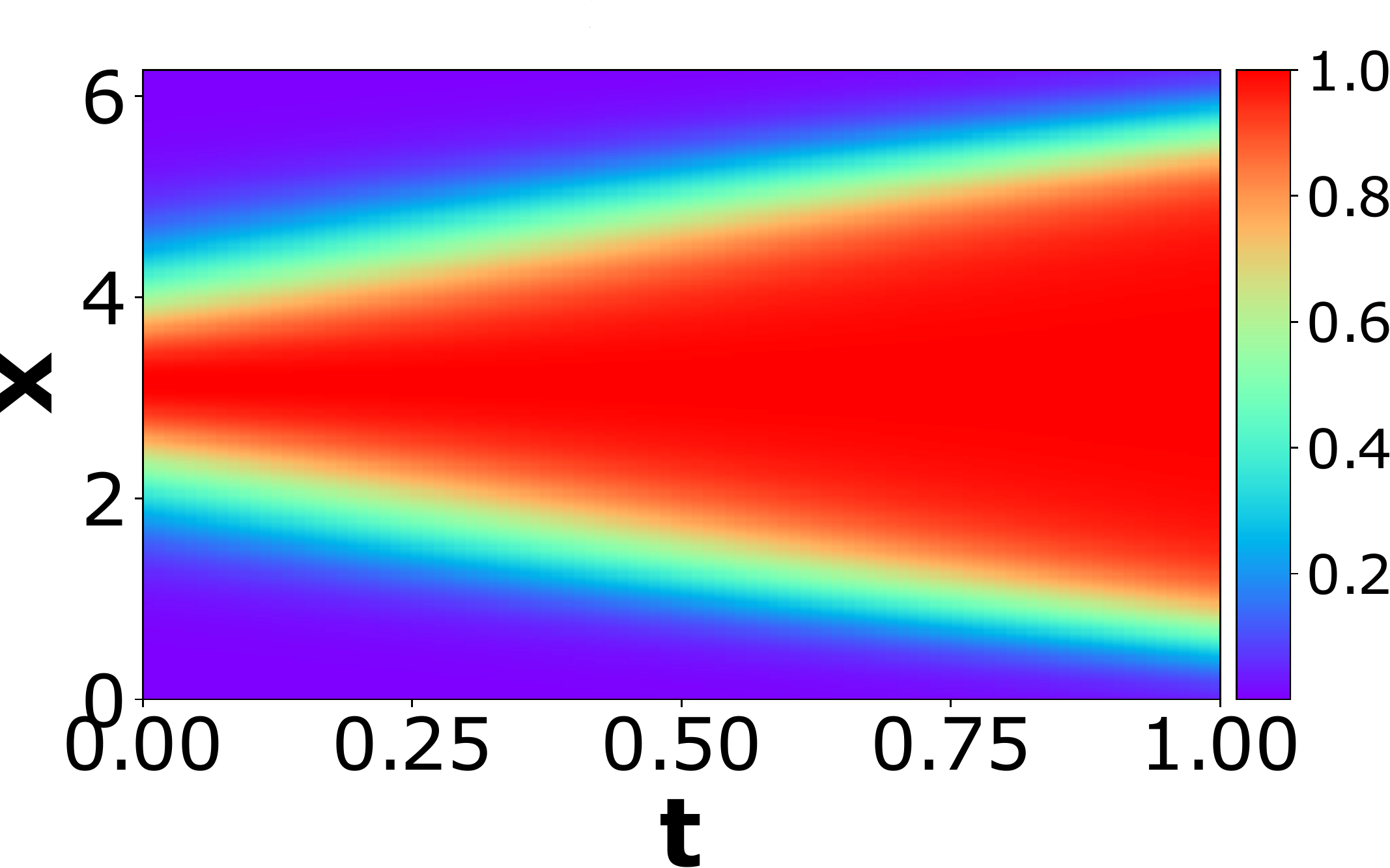}
        \caption{Exact solution.}
        \label{fig:exact_sol}
    \end{subfigure}\hfill
    \begin{subfigure}{0.4\textwidth}
        \centering
        \includegraphics[width=\textwidth]{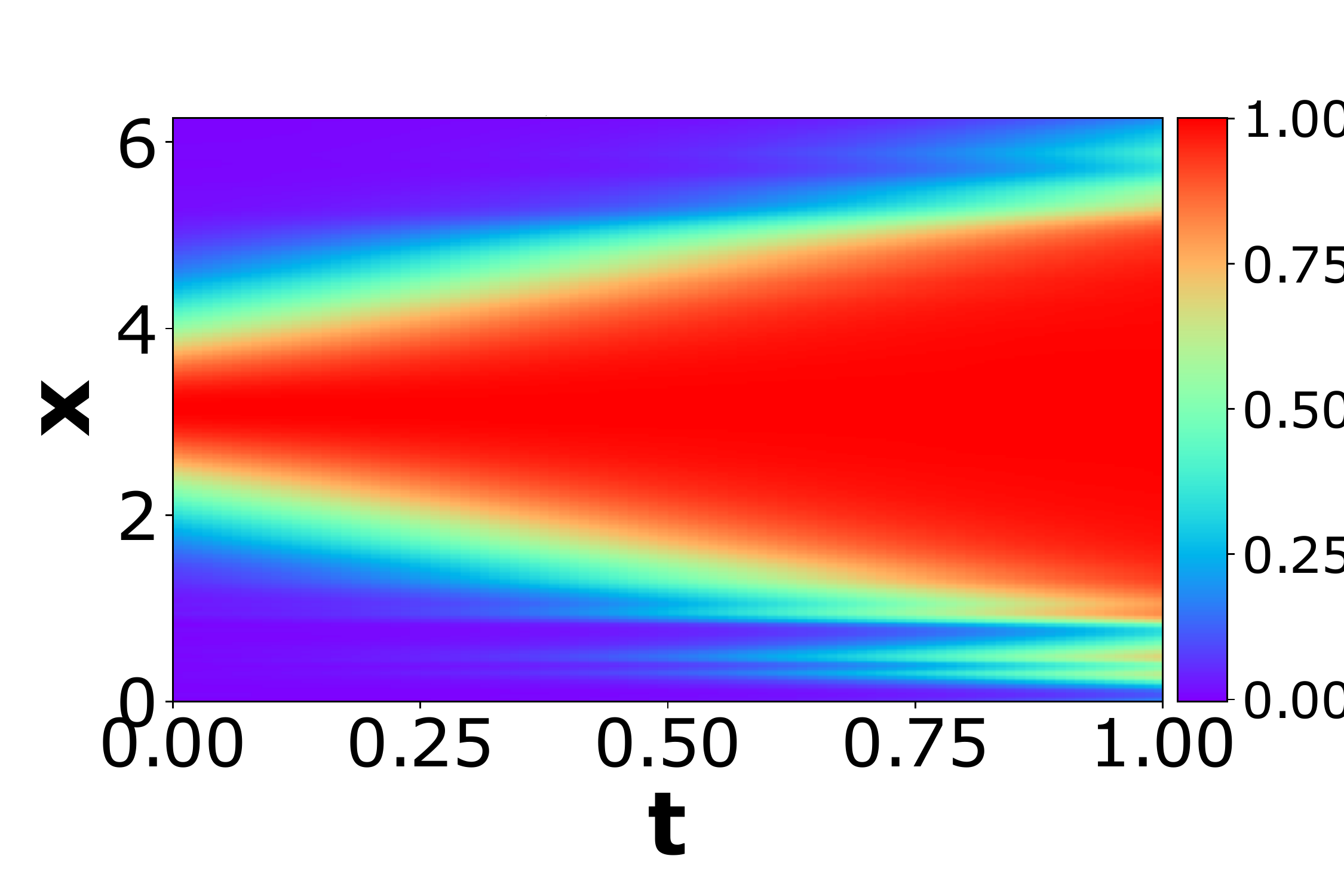}
        \caption{PINNs solution.}
        \label{fig:pinns_sol}
    \end{subfigure}
    \caption{1-D reaction equation with parameter $\rho = 5$ (see \ref{eq:P1.2}).}
    \label{fig:pinns_failure}
\end{figure}

However, sometimes PINNs struggle to learn the ODE/PDE dynamics \cite{wang2021understanding,krishnapriyan2021characterizing,rohrhofer2022understanding,mojgani2022lagrangian} (see Figure \ref{fig:pinns_failure}). Wight \& Zhao \cite{wight2020solving} proposed several techniques to improve the optimization process compared to the original formulation: mini-batch optimization and adaptive sampling of collocation points. Adaptive sampling in time, splits the time interval $[t_0, T] = \cup_{k=0}^K [t_{k-1}, t_k], \ t_K = T$, and solves an equation on the first interval $[t_0, t_1]$, then on $[t_0, t_2]$, and so on up to $[t_0, T]$. Thus, if a solution can be found on a domain $\mathrm{\Omega} \times [t_0, t_{k-1}]$, then the network is pretrained well for the extended domain $\mathrm{\Omega} \times [t_0, t_{k}]$. Krishnapriyan et al. \cite{krishnapriyan2021characterizing} proposed the \textit{seq2seq} approach that splits the domain into smaller subdomains in time and learns the solution on each of the subdomains with a separate network. Thus, both adaptive sampling in time and \textit{seq2seq} are based on the idea of splitting the domain into multiple subdomains, on which solutions can be learned easier. 

As explained in \cite{rohrhofer2022understanding}, improving PINN's solutions by considering small subdomains is possible  because the loss residuals ($\mathcal{L}_{\mathcal{F}}$ term) can be trivially minimized in the vicinity of fixed points, despite corresponding to nonphysical system dynamics that do not satisfy the initial conditions. Therefore, the reduction of the domain improves the convergence of the optimization problem \eqref{eq:pinns_loss} and helps to escape nonphysical solutions. 

Another strategy is to consider transfer learning. Transfer learning is commonly used in computer vision and natural language processing \cite{bengio2012deep,tan2018survey,ruder2019transfer,houlsby2019parameter}. It tries to improve the optimization process by starting with better weight initialization. In PINNs, transfer learning is also successfully used to accelerate the loss convergence \cite{goswami2020transfer,niaki2021physics,chakraborty2021transfer,xu2023transfer}. For instance, Chen et al. \cite{chen2021transfer} apply transfer learning to learn faster different PDEs creating tasks by changing coefficients or source terms in equations. Analogously, curriculum regularization (similar to curriculum learning \cite{bengio2009curriculum}) is proposed in \cite{krishnapriyan2021characterizing} to find good initial weights. 

\paragraph{\bf Our contribution.} %There are many modifications of PINNs in the literature for various tasks, however, there is no method for incremental training of PINNs without expanding the underlying network architecture. Since PINNs parameterize a solution $u(\mathbf{x}, t)$ as an output of a neural network, $\hat{u}(\theta; \mathbf{x}, t) = \mathcal{N}(\theta; \mathbf{x}, t)$, for two PDEs with solutions  $u_1(\mathbf{x}, t)$ and $u_2(\mathbf{x}, t)$ the following condition does not hold: $\hat{u}_1(\theta; \mathbf{x}, t) = \hat{u}_2(\theta; \mathbf{x}, t), \ (\mathbf{x}, t) \in \mathrm{\Omega}\times [t_0, T]$, if the solutions are parameterized with the same the network $\mathcal{N}$. Thus, in our work, 
We propose \textit{incremental PINNs} (iPINNs) and implement this strategy by creating one subnetwork per task such that a complete neural network can learn multiple tasks. Each subnetwork $\mathcal{N}_i$ has its own set of parameters $\theta_i \subset \theta$, and the model is trained sequentially on different tasks. A subnetwork for a new task can overlap with all previous subnetworks, which helps to assimilate the new task. As a result, the network consists of overlapping subnetworks, while the free parameters can be used for future tasks. To illustrate the benefits of the algorithm we consider two problem formulations (Section \ref{sec:problems}). Firstly, we learn a family of equations (e.g., convection) starting from a simple one and incrementally learning new equations from that family. Secondly, we learn a dynamical system that consists of two processes (e.g., reaction-diffusion) by first learning the individual components of the process. Both scenarios demonstrate that the incremental approach enables an iPINN network to learn for cases where regular PINNs fail. %The network uses pretrained parts (subnetworks) to find and train the next (possibly overlapping) subnetwork that cannot be learned well with regular PINN.
To the best of our knowledge, this is the first example where one network can sequentially learn multiple equations without extending its architecture, with the added benefit that performance is significantly improved.

\section{Related work}

Our methodology is based on creating sparse network representations and, similarly to other PINN research, is sensitive to the choice of activation functions. We briefly highlight key related work herein.

\paragraph{\bf Sparse network representation.} Sparse architectures are often advantageous compared to dense ones  \cite{guo2018sparse,ahmad2019can,ye2019adversarial,liao2022achieving}. According to the lottery ticket hypothesis (LTH) \cite{frankle2018lottery}, every randomly initialized network contains a subnetwork that can be trained in isolation to achieve comparable performance as the original network. Based on this observation, the idea of using subnetworks has been adopted in continual learning \cite{mallya2018packnet,sokar2021spacenet,sokaravoiding}. In this paradigm, every subnetwork created is associated with a particular task and used only for this task to make a prediction. One of the approaches to find these tasks-related subnetworks is connections' pruning \cite{lecun1989optimal,hassibi1992second,han2015learning,dong2017learning,dekhovich2021neural} that removes unimportant parameters while exhibiting similar performance.

%\paragraph{Multi-objective loss optimization.} One of the possible issue during training can be loss unbalance in \eqref{eq:pinns_loss}. This problem arises from different gradients' magnitudes of $\mathcal{L}_{\mathcal{F}}$, $\mathcal{L}_{u_0}$ and $\mathcal{L}_{b}$ during training procedure. Some approaches require calculating the weights from the form of an equation \cite{van2022optimally}, while the others are more general and compute the weights from the behavior of loss function or gradients history \cite{wang2021understanding, bischof2021multi}. However, recent research on computer vision and language processing problems shows that multi-tasking optimization approaches in general do not outperform simple scalarization with predetermined fixed weights \cite{xin2022current}. Hence, we do not focus on these approaches in our work.

\paragraph{\bf Choice of the activation function.} There are several studies that investigate how different activation functions affect the performance of neural networks in classification and regression tasks \cite{szandala2021review,jagtap2022important}. It was shown that \texttt{ReLU} \cite{Nair2010RectifiedLU} activation function which can be powerful in classification tasks, in the case of physics-informed machine learning (PIML) regression, may not be the optimal choice. Meanwhile, hyperbolic tangent (\texttt{tanh}) or sine (\texttt{sin}) perform well for PIML. Sinusoidal representation networks (SIRENs) \cite{sitzmann2020implicit} tackle the problem of modeling the signal with fine details. Special weights initialization
scheme combined with \texttt{sin} activation function allows SIREN to learn  complex natural signals. Hence, we use \texttt{sin} activation function in our experiments. In Section \ref{sec:hyperparameters}, we provide the comparison in results between the discussed activation functions.

\section{Problem formulation} \label{sec:problems}

We focus on two scenarios: (1) incremental PINNs learning, where the network sequentially learns several equations from the same family; and (2) learning a combination of multiple equations that create another physical process. To illustrate these cases, we consider one-dimensional convection, reaction and reaction-diffusion problems with periodic boundary conditions.

\subsection{Scenario 1: Equation incremental learning}
We consider the problem of learning the sequence of equations that belong to one family:
\begin{equation}\tag{P1}\label{eq:P1}
    \mathcal{F}_k(u(x, t)) = 0, \quad x \in \mathrm{\Omega}, \ t \in [t_0, T], \ k=1,2,\ldots, K,
\end{equation}

\noindent where $\{\mathcal{F}_k\}_{k=1}^K$ are differential operators from the same family of equations.

\begin{equation*}
\begin{minipage}{.48\linewidth}
    \begin{center}
        \textbf{1-D convection equation}
        \begin{align*}\tag{P1.1}\label{eq:P1.1}
          \frac{\partial u}{\partial t} + \beta_k \frac{\partial u}{\partial x} &= 0,\\
          u(x, 0) &= h_1(x),\\
          u(0, t) &= u(2\pi, t),
        \end{align*}
        where $t \in [0, 1], \ x \in [0, 2\pi], \ \beta_k \in \mathcal{B} \subset \mathbb{N}$.
    \end{center}
\end{minipage}
\vline
\begin{minipage}{.48\linewidth}
    \begin{center}
        \textbf{1-D reaction equation}
        \begin{align*}\tag{P1.2}\label{eq:P1.2}
          \frac{\partial u}{\partial t} - \rho_k u(1-u) &= 0,\\
          u(x, 0) &= h_2(x),\\
          u(0, t) &= u(2\pi, t),
        \end{align*}
        where $t \in [0, 1], \ x \in [0, 2\pi], \ \rho_k \in \mathcal{R} \subset \mathbb{N}$.
    \end{center}
\end{minipage}
\end{equation*}

In this case, every task $k$ is associated with $\mathcal{D}_k = \{(x, t, k):  \ x \in [0, 2\pi], \ t \in [t_0, T], \ k \in \mathbb{N} \}$. Following \cite{krishnapriyan2021characterizing}, we take $h_1(x) = \sin{x}$ and $h_2(x) = e^{-\frac{(x-\pi)^2}{2(\pi/4)^2}}$.

\subsection{Scenario 2: Combination of multiple equations}

We also consider the case when a dynamic process consists of multiple components. Let us consider the reaction-diffusion equation:

\begin{align*} \tag{P2}\label{eq:P2}
    \frac{\partial u}{\partial t} - \nu \frac{\partial^2 u}{\partial x^2} - \rho u(1-u) &= 0, \\
    u(x, 0) &= h_2(x),\\
    u(0, t) &= u(2\pi, t),
\end{align*}

\noindent where $t \in [0, 1], \ x \in [0, 2\pi], \ \nu, \rho > 0$. This process consists of two parts: reaction term $(\nu = 0)$: $-\rho u(1-u)$ and diffusion term $(\rho = 0)$: $-\nu \frac{\partial^2 u}{\partial x^2}$. Therefore, we construct one task as the reaction, another one as the diffusion, and the final one as the reaction-diffusion. We can change the order of the reaction tasks and diffusion tasks to show the robustness of incremental learning. The reaction-diffusion task should be the last one since our goal is first to learn the components of the system and only then the full system. 

Considering these two problems, we want to show that better generalization can be achieved by pretraining the network with simpler related problems rather than by dividing the domain into smaller subdomains. In the following section, we show how one network can incrementally learn different equations without catastrophic forgetting.

\section{Methodology}

\begin{figure}[ht]
    \centering
    \includegraphics[width=0.9\textwidth]{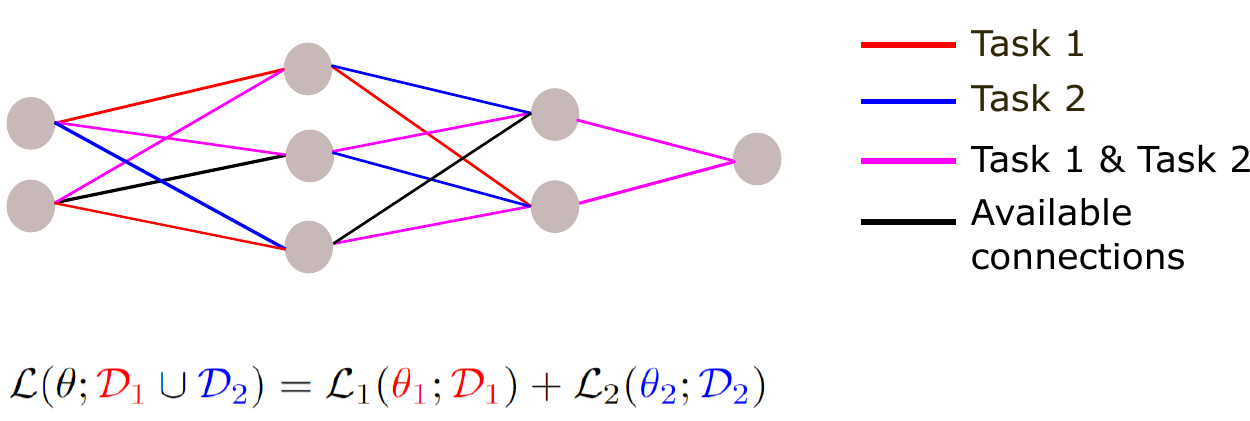}
    \caption{An example of iPINNs with two PDEs: every subnetwork corresponds to only one task (PDE).}
    \label{fig:ipinns}
\end{figure}

The proposed method needs to be applicable to both types of problems \ref{eq:P1} and \ref{eq:P2}. However, these problems cannot be solved by one network with the same output head for all $K$ tasks, since $\mathcal{F}_i(u(x, t)) \ne \mathcal{F}_j(u(x, t))$ for $i \ne j$ and $x \in \mathrm{\Omega}, \ t \in [t_0, T]$. Therefore, we propose iPINNs -- an incremental learning algorithm that focuses on learning task-specific subnetworks $\mathcal{N}_1, \mathcal{N}_2,...,\mathcal{N}_K$ for each task $k=1,2,\ldots, K$. To create these subnetworks, we use an iterative pruning algorithm NNrelief \cite{dekhovich2021neural}. This pruning approach uses input data to estimate the contribution of every connection to the neuron in the pretrained network and delete the least important ones. However, in principle, any connections pruning algorithm or any other approach that is able to find and train sparse network representations is suitable.

iPINN trains task-related subnetworks with pruning, allowing the subnetworks to overlap on some connections. This way the method provides knowledge sharing between the subnetworks. These overlaps are updated with respect to all tasks that are assigned to a particular connection. Let us denote the loss of each task $\mathcal{D}_j$ as $\mathcal{L}_j = \mathcal{L}(\theta_j; \mathcal{D}_j)$, where $\theta_j$ is the parameter vector for task $\mathcal{D}_j$, $1 \le j \le k$. Then the total loss and its gradient with respect to a parameter $w$  can be written as:

\begin{align}
    \mathcal{L} &= \sum_{j=1}^k \mathcal{L}_j, \label{eq:total_loss}\\
    \frac{\partial \mathcal{L}}{\partial w} &= \sum_{j=1}^k \frac{\partial \mathcal{L}_j}{\partial w} = \sum_{j: \ w \in \mathcal{N}_j} \frac{\partial \mathcal{L}_j}{\partial w} \label{eq:total_grad},
\end{align}

because if $w \not \in \mathcal{N}_j$ , then $\frac{\partial \mathcal{L}_j}{\partial w} = 0$. The pseudocode of the algorithm is shown as follows:

\begin{minipage}{0.9\textwidth}
    \begin{algorithm}[H]
      \begin{algorithmic}[1]
        \Require neural network $\mathcal{N}$, training datasets $\mathcal{D}_k \ (k=1,2, \dots, K) $, training hyperparameters, pruning hyperparameters ($num\_iters$).
        \For {$k = 1, 2, \ldots, K$}
            \State{$\mathcal{N}_k \gets \mathcal{N}$ } \Comment{set full network as a subnetwork}
            \State{Train $\mathcal{N}_1, \mathcal{N}_2, \ldots, \mathcal{N}_k$ on tasks $\mathcal{D}_1, \mathcal{D}_2, \ldots, \mathcal{D}_k$ using Eq. \ref{eq:total_grad}.}
            \For {$it = 1,2, \ldots, num\_iters$}    \Comment{repeat pruning}
                \State{$\mathcal{N}_k \gets Pruning(\mathcal{N}_k, \mathcal{D}_k)$}    \Comment{reduce unimportant connections}
                \State{Retrain subnetworks $\mathcal{N}_1, \mathcal{N}_2, \ldots, \mathcal{N}_k$ on tasks $\mathcal{D}_1, \mathcal{D}_2, \ldots, \mathcal{D}_k$ using Eq. \ref{eq:total_grad}.}
            \EndFor
        \EndFor
      \end{algorithmic}
    \caption{PINN incremental learning}
    \label{alg:alg1}
    \end{algorithm}
\end{minipage}

\vskip 10pt

The main advantage of the proposed approach is that a neural network learns \textit{all} tasks (subdomains or equations) that were given during training and not only the last one. This is achieved by constantly replaying old data. In the next section, we experimentally show that pretrained parts of the network help to improve the convergence process.

\section{Numerical experiments}

Our findings illustrate the advantage of the Algorithm over regular PINNs \cite{raissi2019physics}. The Algorithm allows the network to learn multiple equations (\ref{eq:P1}) from the same family. Furthermore, by starting with simpler tasks, the network can learn more complex ones that cannot be learned separately.

\paragraph{\bf Experiments setup.} Let us start by examining the proposed algorithms on the convection and reaction equations with periodic boundary conditions (\ref{eq:P1}). Following the setup in \cite{krishnapriyan2021characterizing}, we use a four-layer neural network with 50 neurons per layer. We use 1000 randomly selected collocation points on every time interval between 0 and 1 for $\mathcal{L}_{\mathcal{F}}$. The Adam optimizer \cite{kingma2014adam} is used to train the model.

To evaluate the performance of the algorithms we compare the final error after the last task. In addition, following continual learning literature \cite{lopez2017gradient}, we compare backward and forward transfer metrics. Let us denote the test set as $\mathcal{D}^{test} = \{(x^i, t^i, l): x^i \in [0, 2\pi], \ t^i \in [0, 1], \ l \text{ is the task-ID} \}$, the solution of the equation at the point $(x^i, t^i, l)$ as $\mathbf{u}_{l,k}^i = u_{l,k}^i(x^i, t^i)$, and $\mathbf{\hat{u}}_{l,k}^i$ is a prediction of the model at point $(x^i, t^i, l)$ after task $\mathcal{D}_k$ is learned. Relative and absolute errors are denoted as $r_{l,k}$ and $\varepsilon_{l, k}$, respectively, as they are calculated for task $l$ after task $k$ is learned ($l \le k$).
\begin{align}
    \text{Relative error:} \quad r_{l, k} &= \frac{1}{N}\frac{\lvert\lvert \mathbf{u}_{l} - \mathbf{\hat{u}}_{l, k} \rvert\rvert_2}{\lvert\lvert \mathbf{u}_l \rvert\rvert_2} \times 100\%,\\
    \text{Absolute error:} \quad \varepsilon_{l, k} &= \frac{1}{N} \sum_{i=1}^N \lvert \mathbf{u}_l^i - \mathbf{\hat{u}}_{l,k}^i \rvert,\\
    \text{Backward Transfer:} \quad \text{BWT} &= \frac{1}{k-1} \sum_{l=1}^{k-1} \varepsilon_{l,k} - \varepsilon_{l,l} \ \text{ or } \\  \text{BWT} &= \frac{1}{k-1} \sum_{l=1}^{k-1} r_{l,k} - r_{l,l}
\end{align} 

\subsection{Results}\label{sec:results}

%In this section, we show how we achieve better generalization than regular PINNs by learning sequentially the same equation with varying parameters.
Table \ref{tab:reaction1-5} presents the results after all reaction equations are learned varying $\rho$ from 1 to 5. Figure \ref{fig:reaction1-5} shows the error history for every equation after incremental steps. The Table summarizes the performance improvement of iPINNs compared to regular PINNs, exhibiting negligible error for all values of $\rho$,  which is especially relevant for cases when $\rho$ is larger. Moreover, iPINNs provide negative BWT which means that previous subnetworks help to learn the following ones.

\begin{table}[ht!]
    \begin{minipage}{0.45\linewidth}
    \caption{Final error and forgetting after all reaction equations are learned.}
        \centering
            \begin{tabular}{lccc}
                \toprule
                 & & regular PINN & iPINN\\
                \midrule
                \multirow{2}{*}{$\rho = 1$} & abs. err & $1.09 \times 10^{-3}$ & $\mathbf{1.5 \times 10^{-4}}$\\
                                            & rel. err & 0.263\% & $\mathbf{0.039}$\% \\
                \midrule
                \multirow{2}{*}{$\rho = 2$} & abs. err & $1.97 \times 10^{-3}$ & $\mathbf{2.5 \times 10^{-4}}$\\
                                            & rel. err & 0.479\% & $\mathbf{0.070\%}$ \\
                \midrule
                \multirow{2}{*}{$\rho = 3$} & abs. err & $6.72 \times 10^{-3}$ & $\mathbf{6.1 \times 10^{-4}}$\\
                                            & rel. err & 2.05\% & $\mathbf{0.210}\%$ \\
                \midrule
                \multirow{2}{*}{$\rho = 4$} & abs. err & $1.13 \times 10^{-2}$ & $\mathbf{1.18 \times 10^{-3}}$\\
                                            & rel. err & 3.68\% & $\mathbf{0.458\%}$ \\
                \midrule
                \multirow{2}{*}{$\rho = 5$} & abs. err & $5.04 \times 10^{-2}$ & $\mathbf{1.91 \times 10^{-3}}$\\
                                            & rel. err & 12.19\% & $\mathbf{0.763\%}$ \\
                \midrule\midrule
                \multirow{2}{*}{BWT} & abs. err & N/A & -$3.8 \times 10^{-4}$\\
                                     & rel. err & N/A & -0.112\%\\
            \bottomrule
            \end{tabular}
            \label{tab:reaction1-5}
    \end{minipage}\hfill
    \begin{minipage}{0.45\linewidth}
    \centering
    \includegraphics[width=\textwidth]{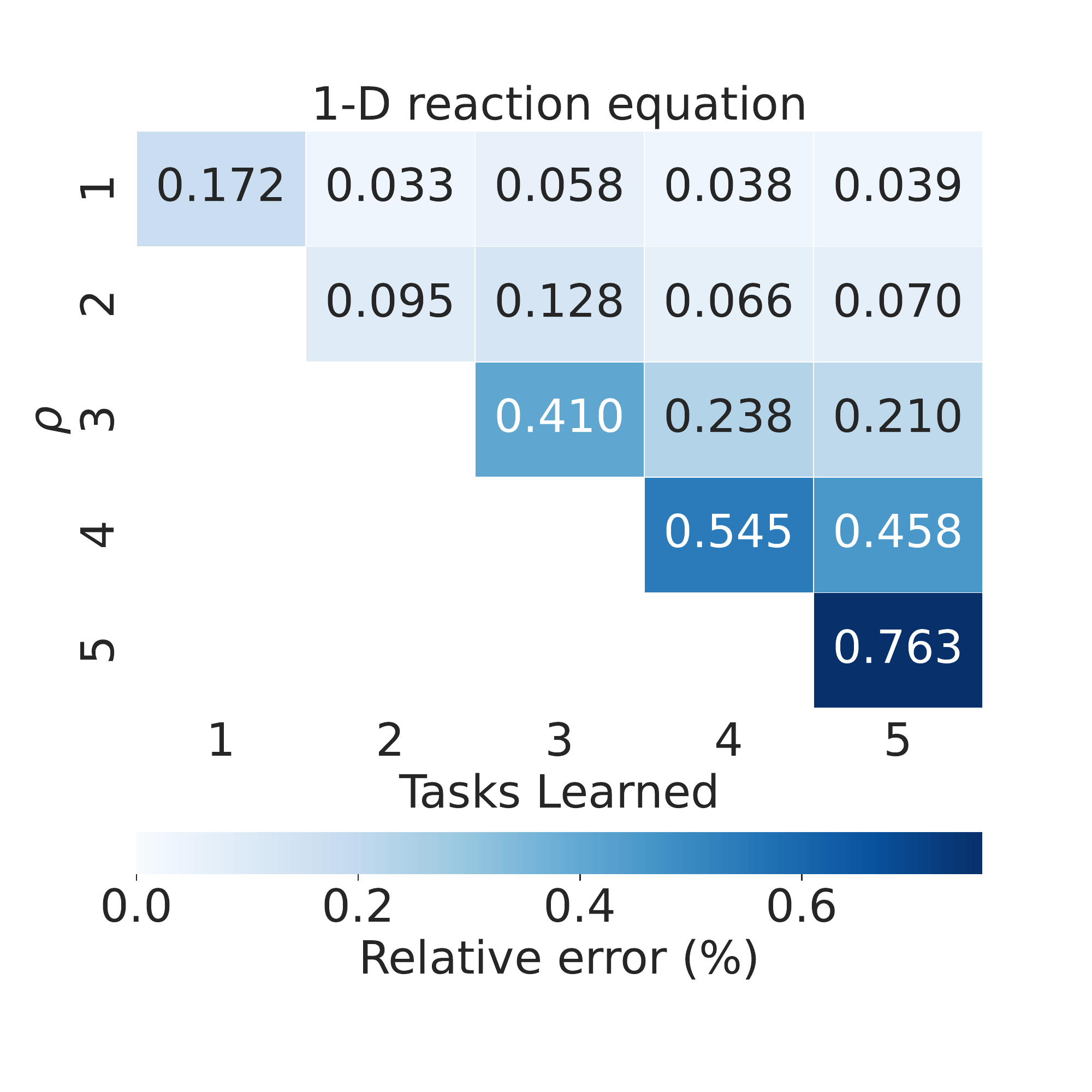}
    \captionof{figure}{Relative error history for reaction eqations ($\alpha=0.99$). Every row shows the error after a new task is learned.}
    \label{fig:reaction1-5}
    \end{minipage}
\end{table}

Similarly, we observe for the convection equation the same learning behaviour. By learning incrementally the sequence of convection equations, we achieve much lower absolute and relative errors for the equations that are more difficult to learn ($\beta =30, 40$). In Table \ref{tab:convection1-40} we show final errors at the end of training, and Figure \ref{fig:convection1-40} shows the absolute error history for each equation.

\begin{table}[ht!]
    \begin{minipage}{0.45\linewidth}
    \caption{Final error and forgetting after all convection equations are learned.}
        \centering
            \begin{tabular}{lccc}
                \toprule
                 & & regular PINN & iPINN\\
                \midrule
                \multirow{2}{*}{$\beta = 1$} & abs. err & $\mathbf{2.3 \times 10^{-4}}$  & $4.2 \times 10^{-4}$ \\
                                            & rel. err & $\mathbf{0.042\%}$& 0.074\%\\
                \midrule
                \multirow{2}{*}{$\beta = 10$} & abs. err & $1.3 \times 10^{-3}$ & $\mathbf{5.0 \times 10^{-4}}$\\
                                            & rel. err & 0.222\% & $\mathbf{0.087\%}$\\
                \midrule
                \multirow{2}{*}{$\beta = 20$} & abs. err & $1.9 \times 10^{-3}$ & $\mathbf{1.66 \times 10^{-3}}$\\
                                            & rel. err & 0.339\% & $\mathbf{0.288\%}$ \\
                \midrule
                \multirow{2}{*}{$\beta = 30$} & abs. err & $2.2 \times 10^{-1}$  & $\mathbf{1.44 \times 10^{-3}}$\\
                                            & rel. err & 3.957\% & $\mathbf{0.246\%}$ \\
                \midrule
                \multirow{2}{*}{$\beta = 40$} & abs. err & $2.3 \times 10^{-1}$ & $\mathbf{6.02 \times 10^{-3}}$\\
                                            & rel. err & 37.4\% & $\mathbf{1.139\%}$ \\
                \midrule\midrule
                \multirow{2}{*}{BWT} & abs. err & N/A & $1.8 \times 10^{-4}$\\
                                     & rel. err & N/A & 0.0280\%\\
            \bottomrule
            \end{tabular}
            \label{tab:convection1-40}
    \end{minipage}\hfill
    \begin{minipage}{0.45\linewidth}
    \centering
    \includegraphics[width=\textwidth]{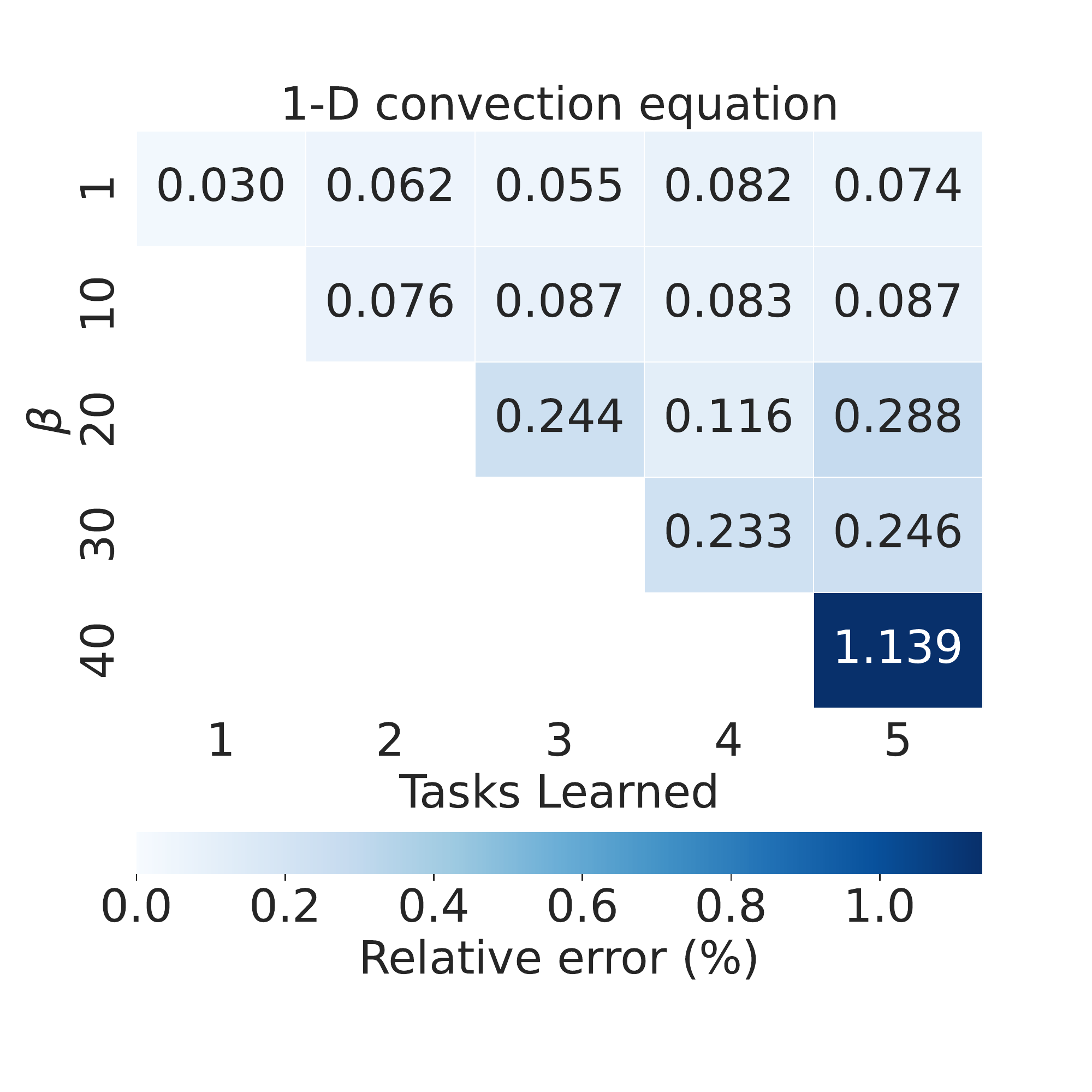}
    \captionof{figure}{Relative error history for convection eqations ($\alpha=0.95$). Every row shows the error after a new task is learned.}
    \label{fig:convection1-40}
    \end{minipage}
\end{table}

\begin{figure}[ht!]
    \begin{subfigure}{\textwidth}
        \centering
        \begin{minipage}{0.32\textwidth}
            \centering
            \includegraphics[width=\linewidth]{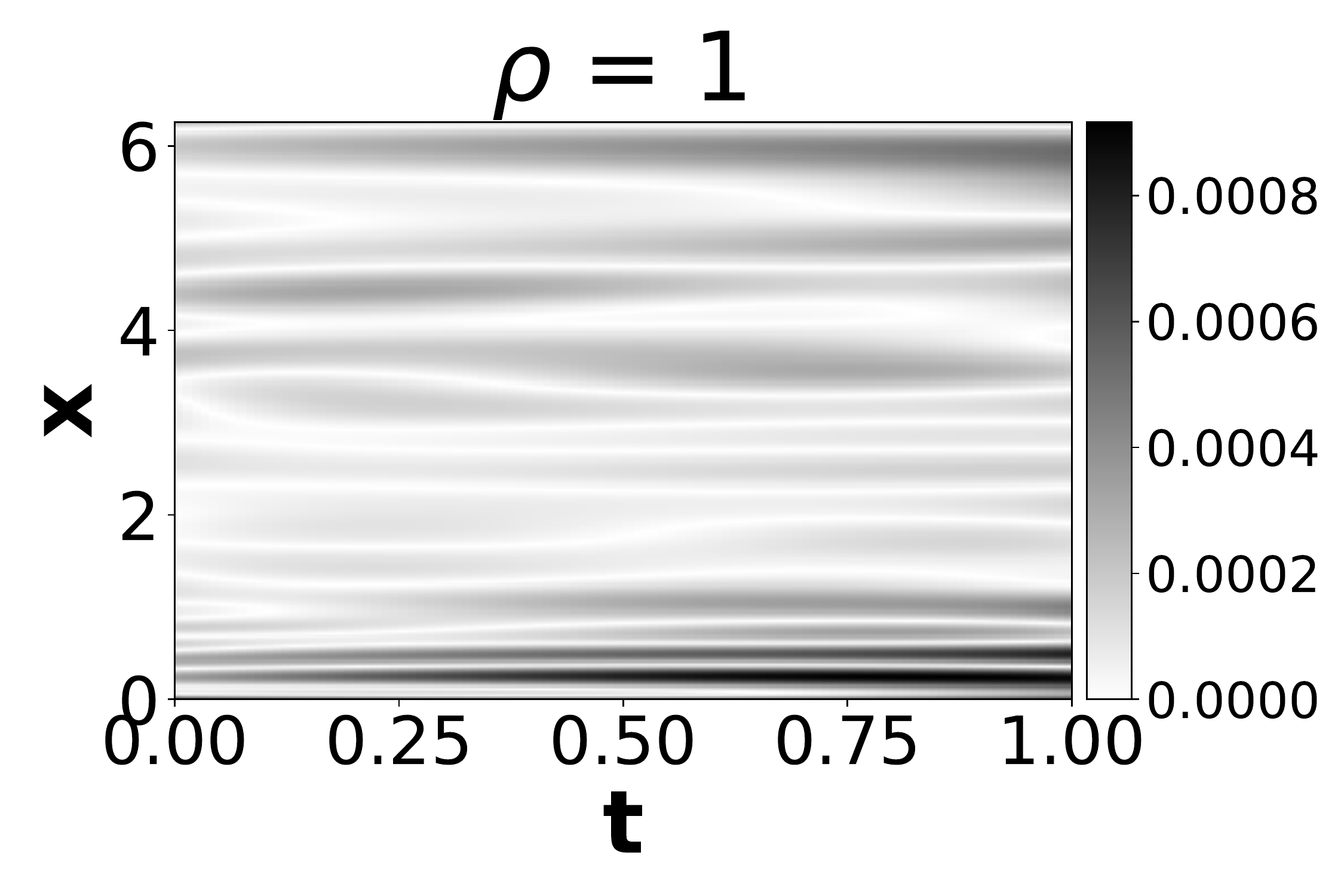}
        \end{minipage}
        \begin{minipage}{0.32\textwidth}
            \centering
            \includegraphics[width=\linewidth]{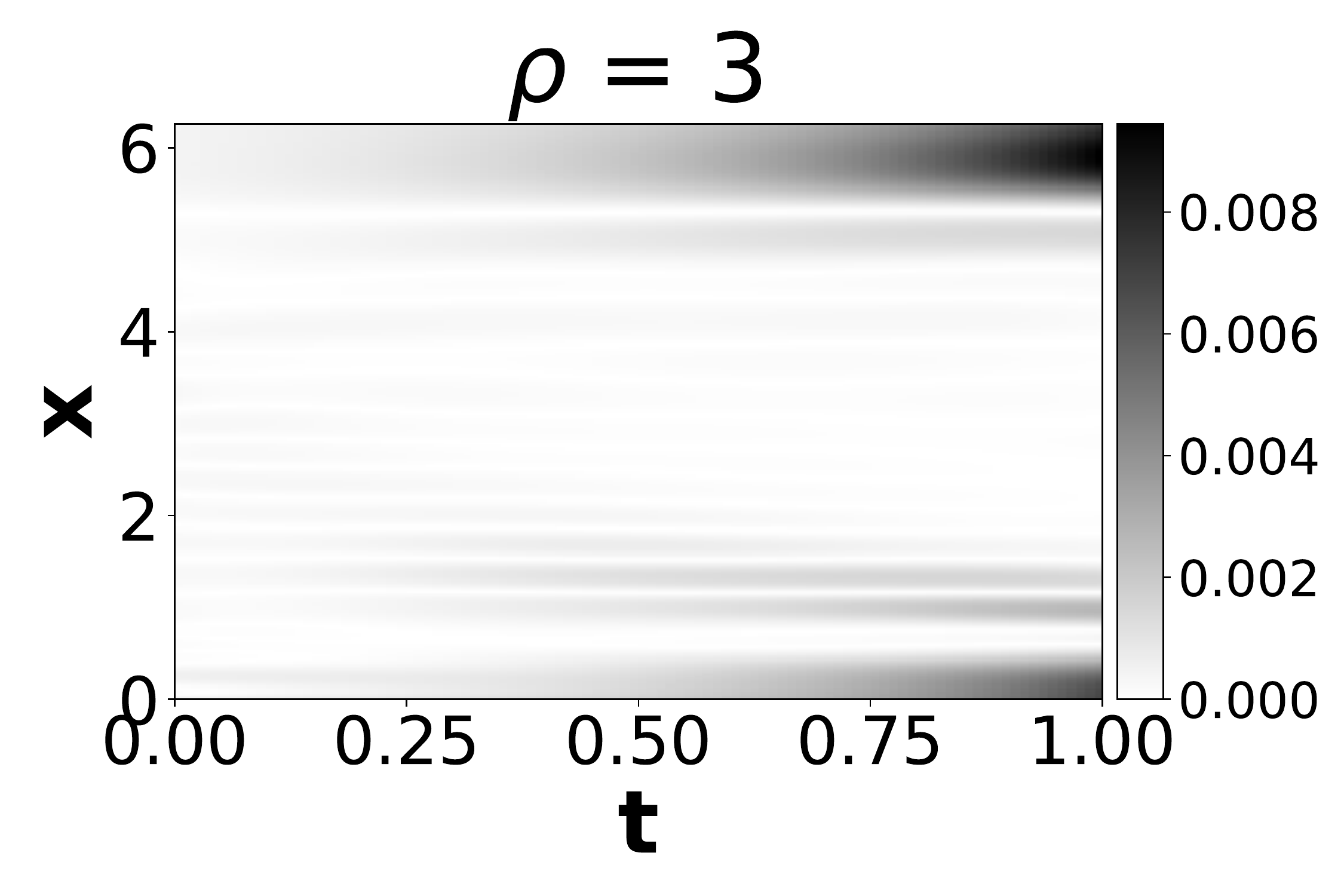}
        \end{minipage}
        \begin{minipage}{0.32\textwidth}
            \centering
           \includegraphics[width=\linewidth]{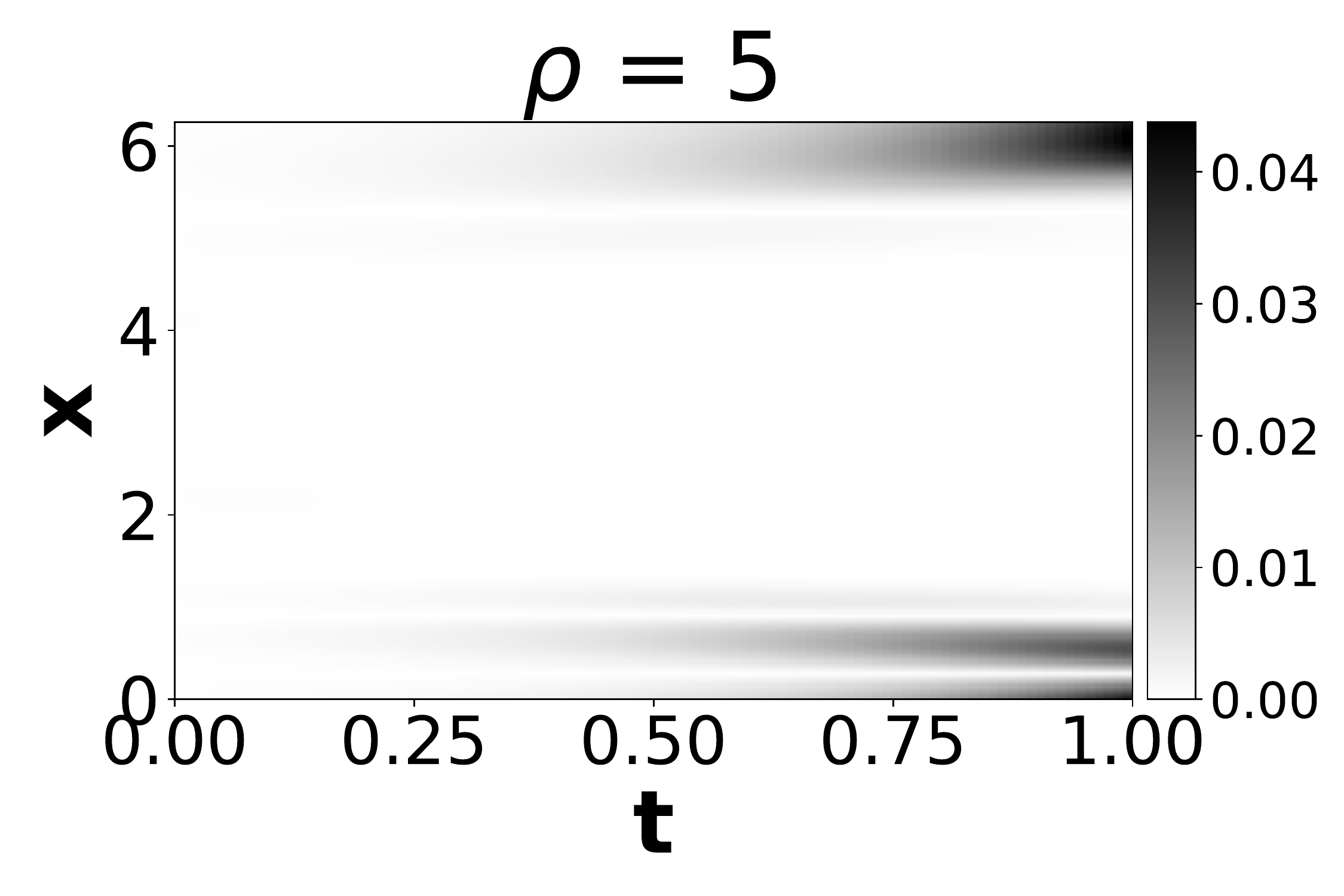}
        \end{minipage}
        \caption{1-D reaction equation (absolute error).}
    \end{subfigure}
    \begin{subfigure}{\textwidth}
        \centering
        \begin{minipage}{0.32\textwidth}
            \centering
            \includegraphics[width=\linewidth]{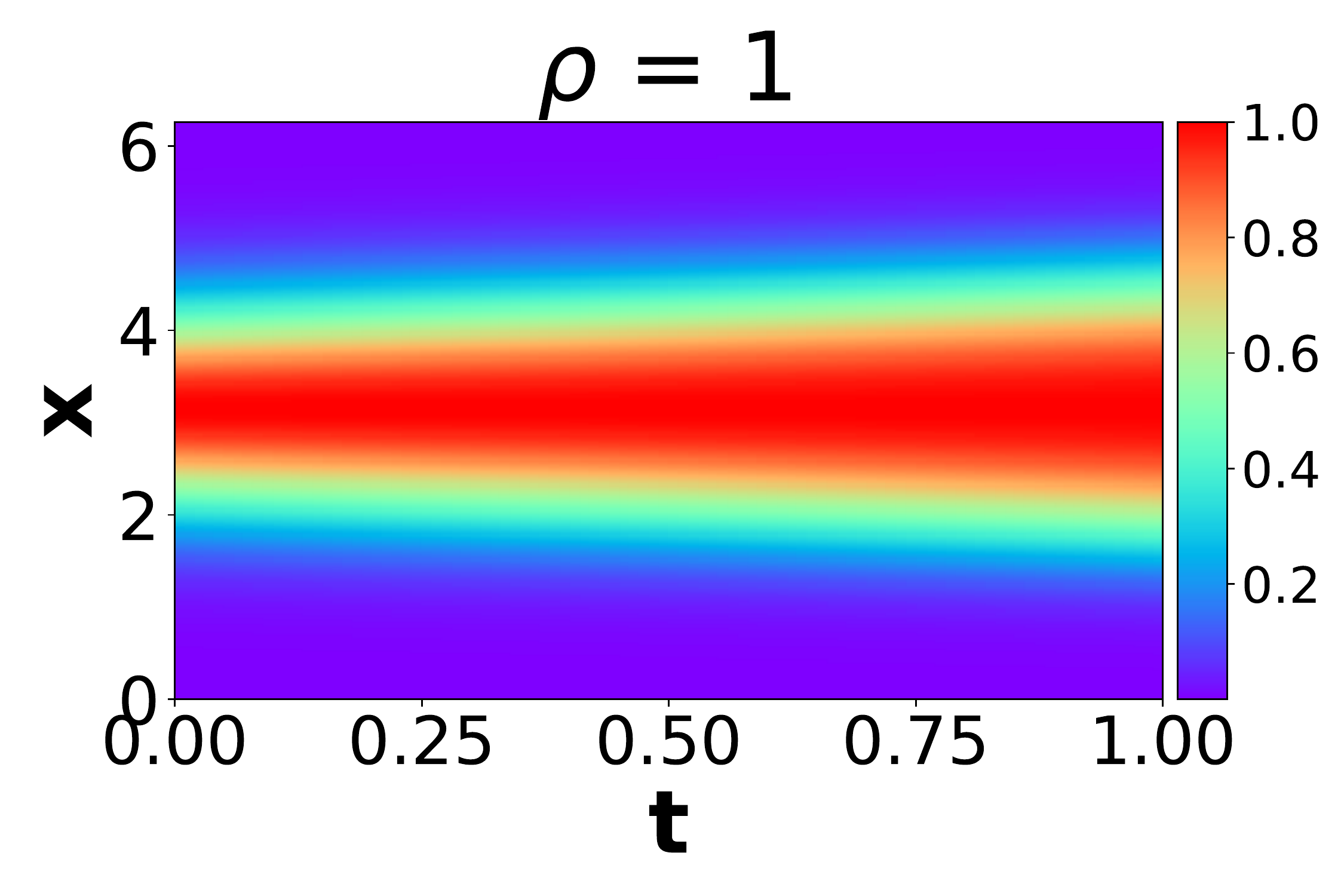}
        \end{minipage}
        \begin{minipage}{0.32\textwidth}
            \centering
            \includegraphics[width=\linewidth]{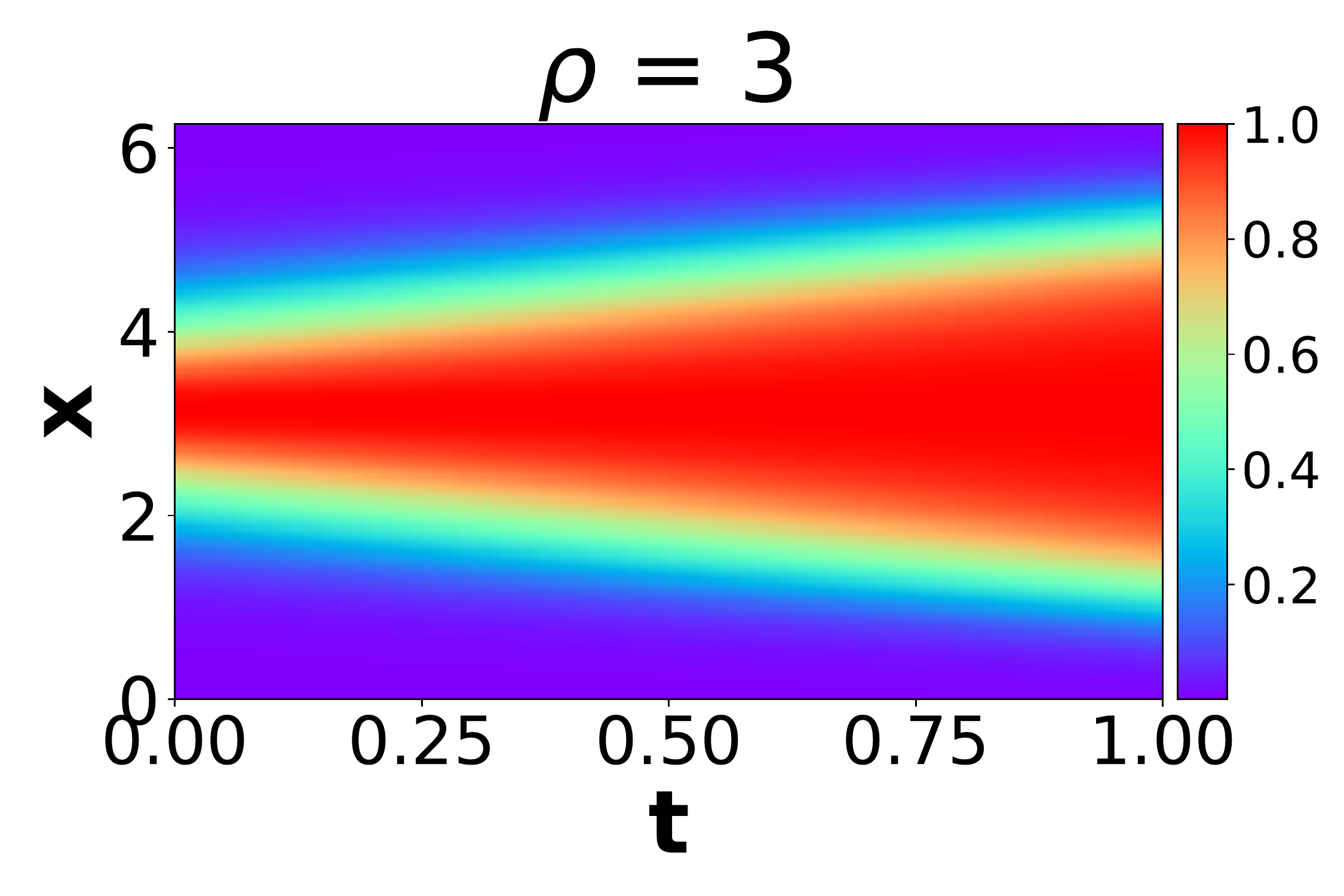}
       \end{minipage}
        \begin{minipage}{0.32\textwidth}
            \centering
            \includegraphics[width=\linewidth]{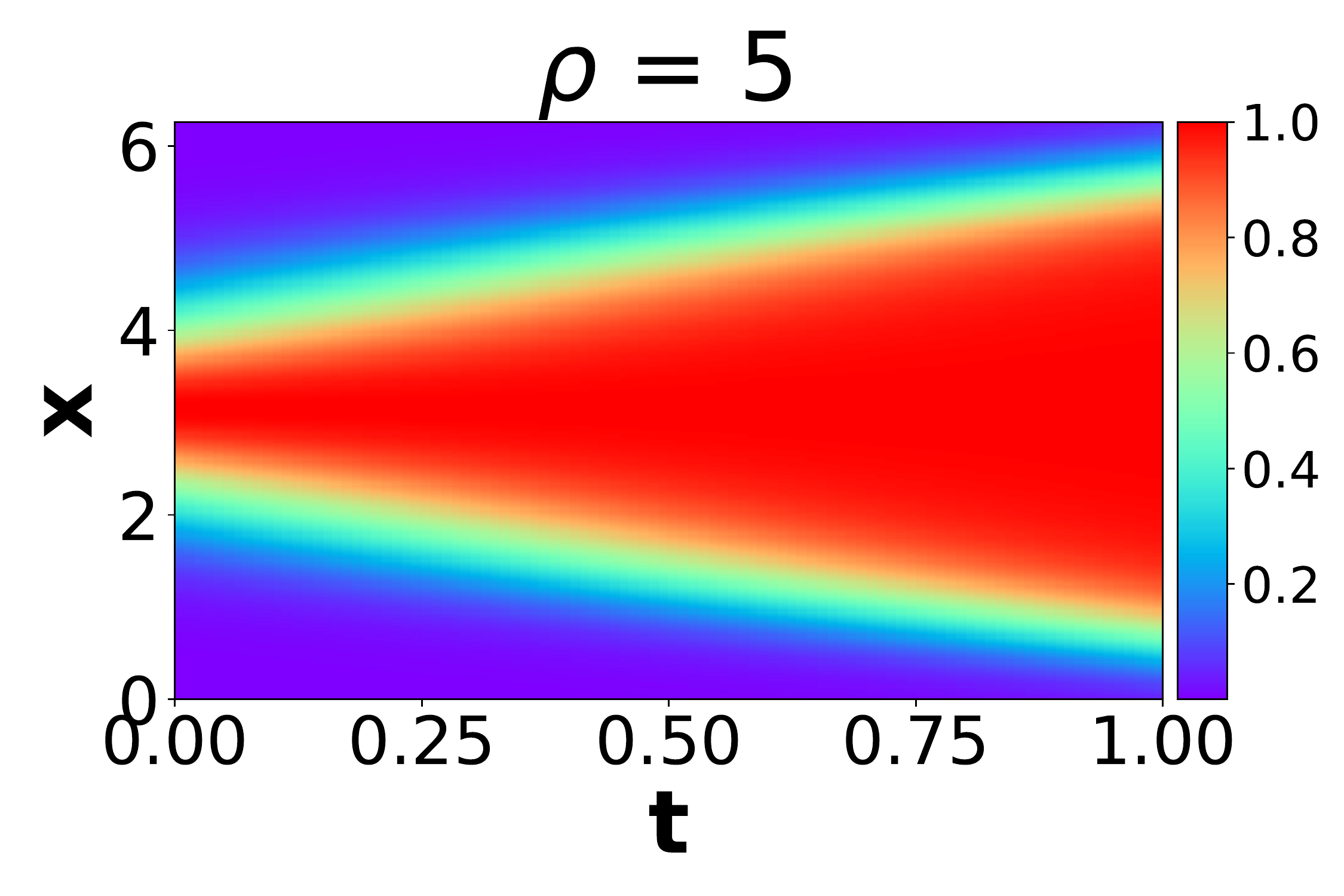}
        \end{minipage}
        \caption{1-D reaction equation (exact solution).}
    \end{subfigure}
    \caption{iPINNs on 1-D reaction equation.}
    \label{fig:error_reaction}
\end{figure}

\begin{figure}[ht!]
    \begin{subfigure}{\textwidth}
        \centering
        \begin{minipage}{0.32\textwidth}
            \includegraphics[width=\textwidth]{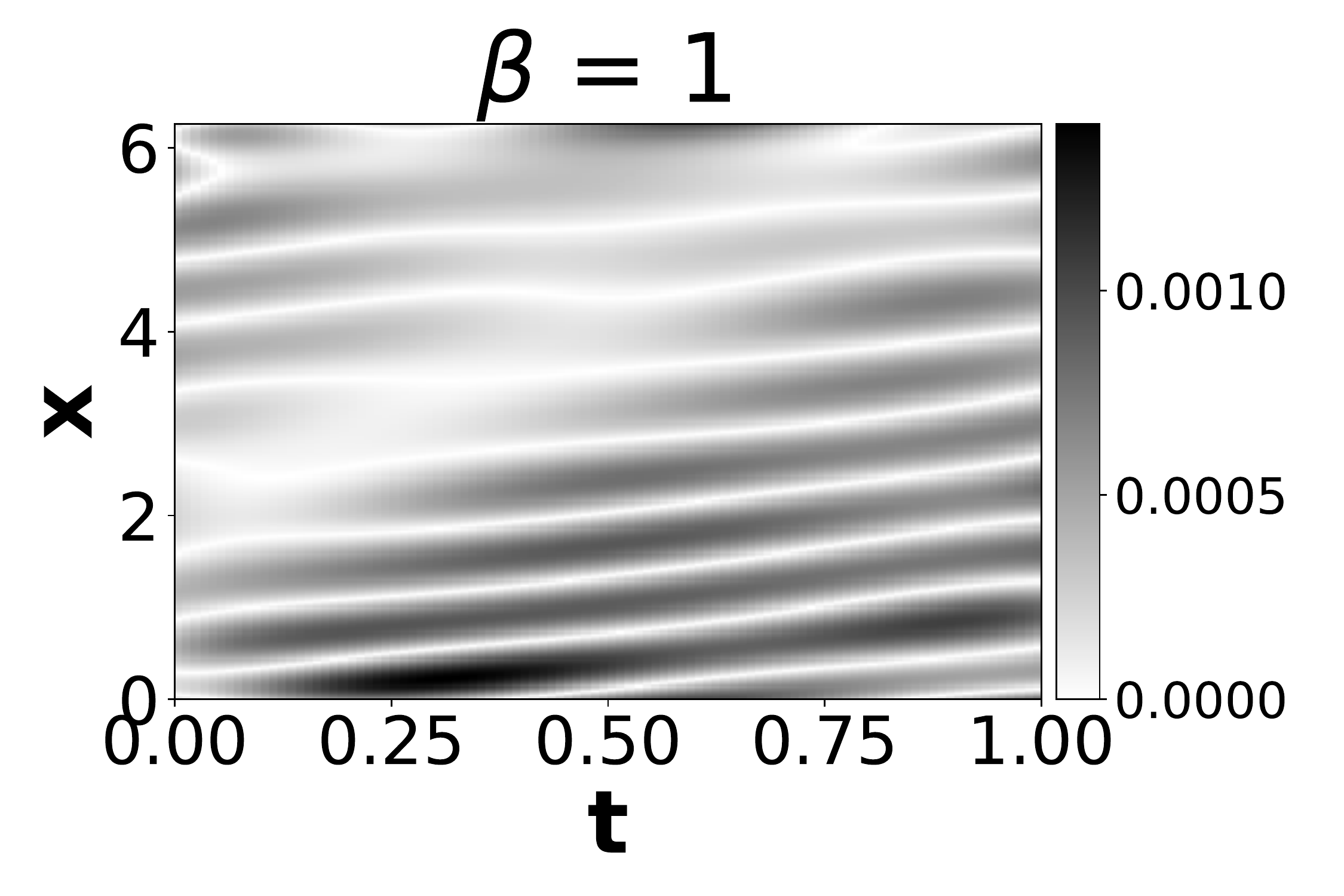}
        \end{minipage}
        \begin{minipage}{0.32\textwidth}
            \includegraphics[width=\textwidth]{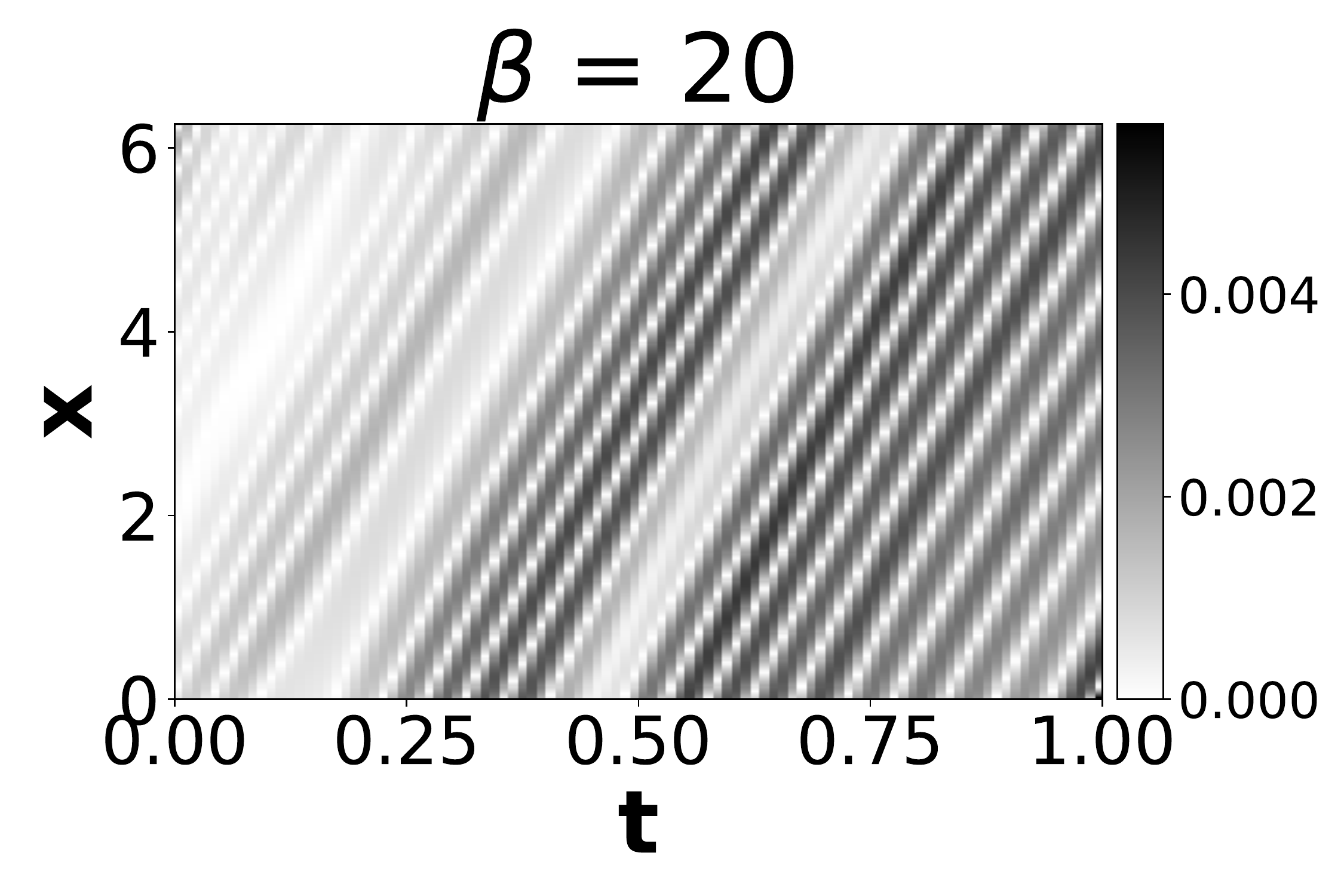}
        \end{minipage} 
        \begin{minipage}{0.32\textwidth}  
            \includegraphics[width=\textwidth]{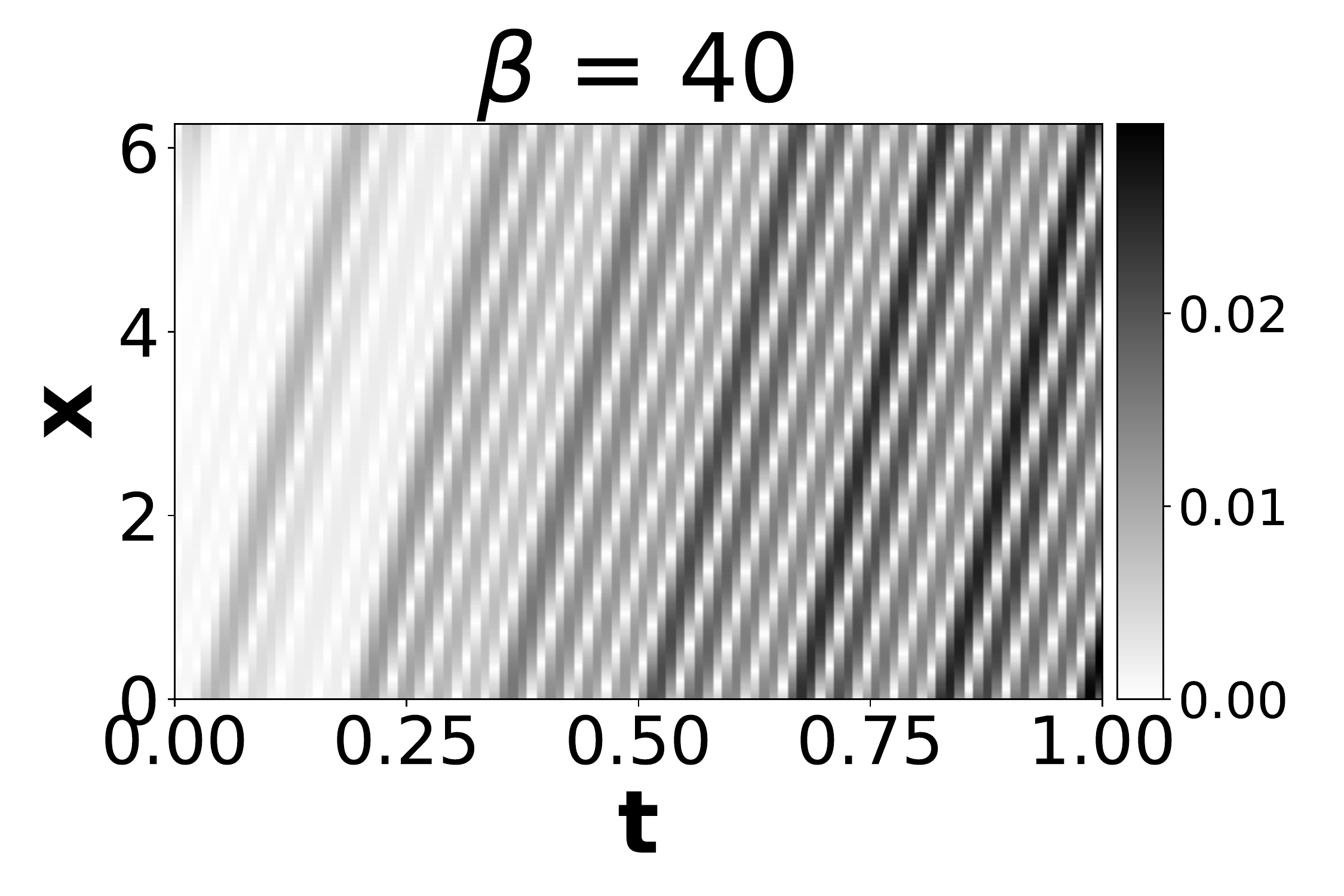} 
        \end{minipage}
        \caption{Absolute error.}
    \end{subfigure}
    \begin{subfigure}{\textwidth}
        \centering
        \begin{minipage}{0.32\textwidth}
            \includegraphics[width=\textwidth]{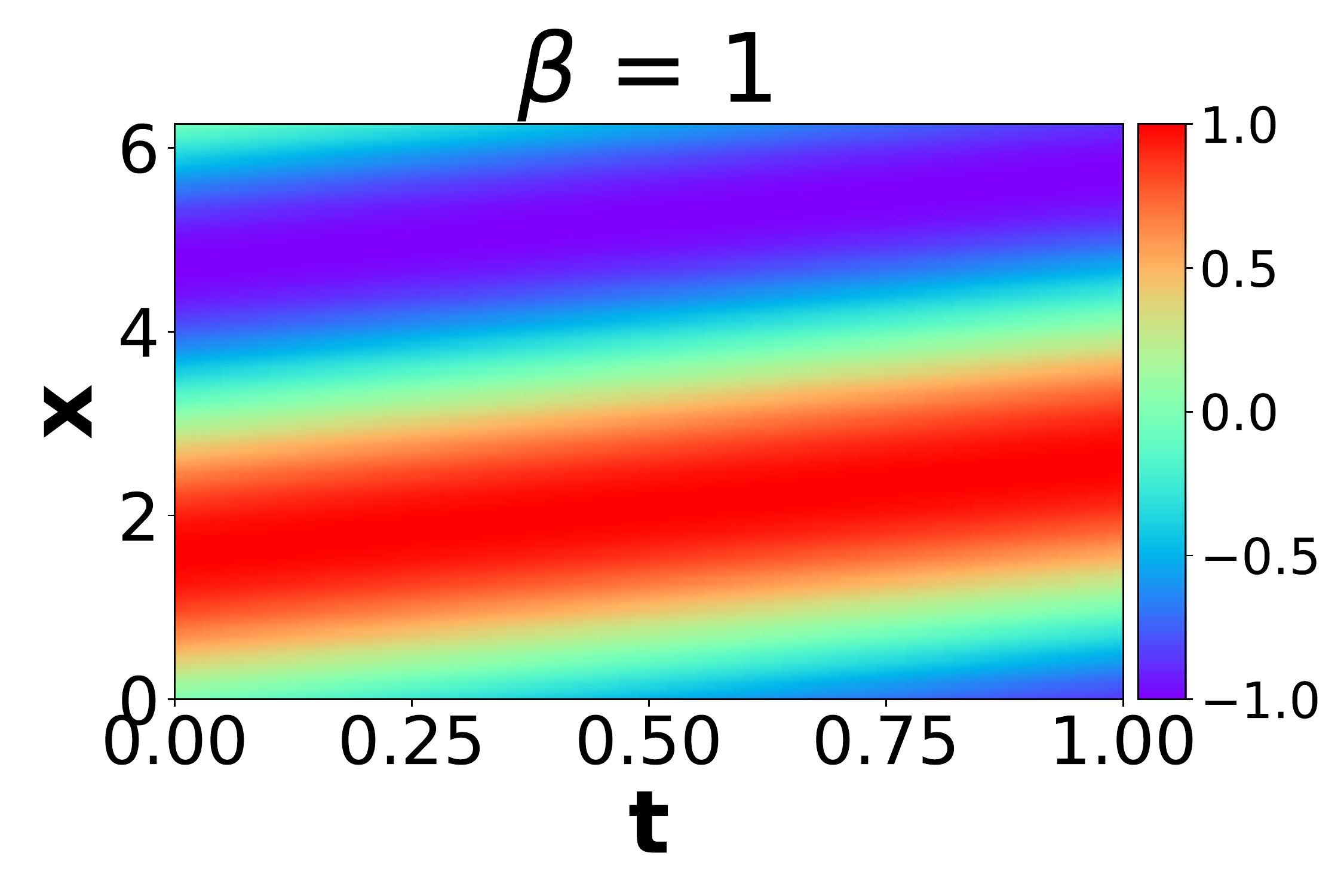}
        \end{minipage}
        \begin{minipage}{0.32\textwidth}
            \includegraphics[width=\textwidth]{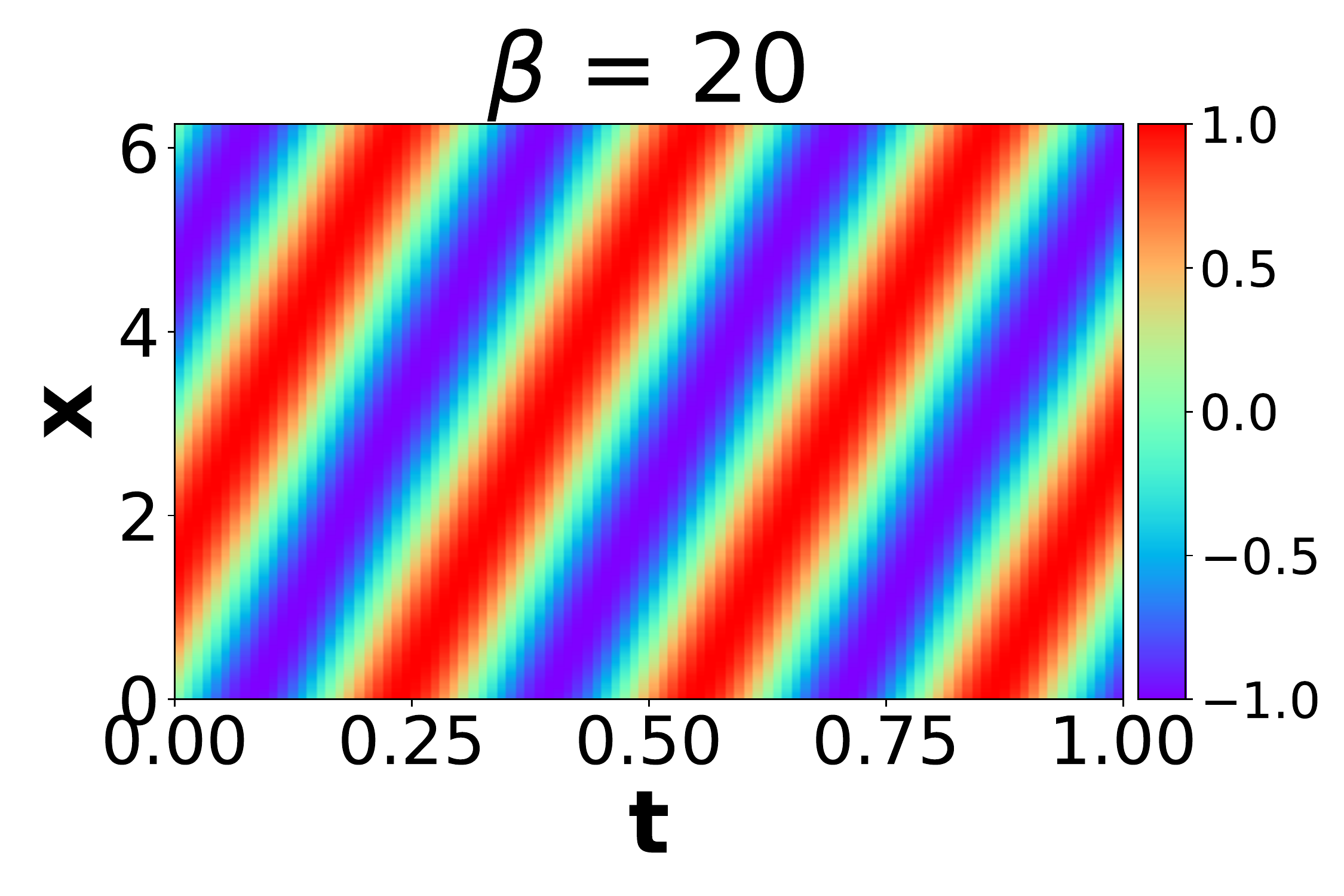}
        \end{minipage}   
        \begin{minipage}{0.32\textwidth}  
            \includegraphics[width=\textwidth]{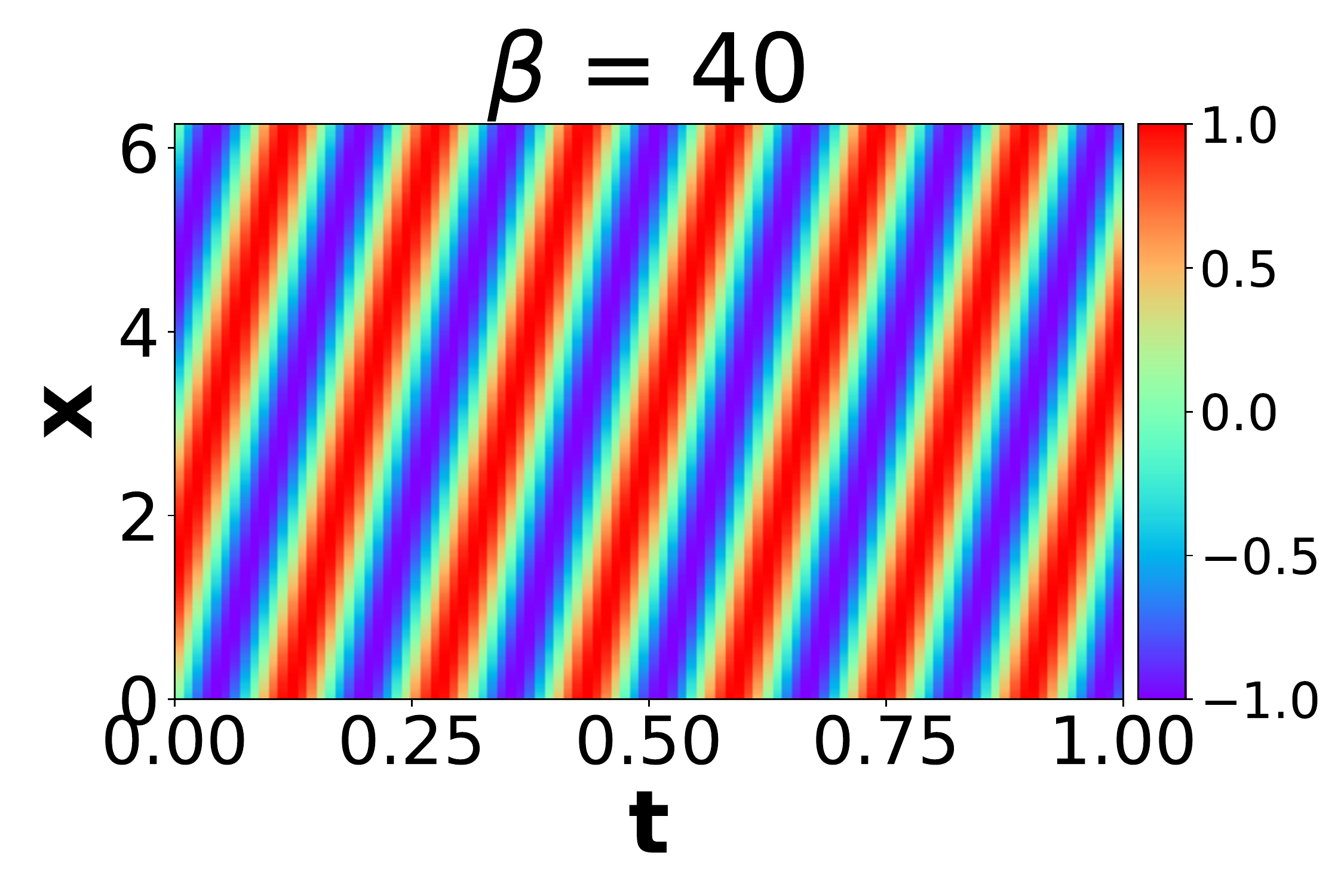} 
        \end{minipage}
        \caption{Exact solution.}
    \end{subfigure}
    \caption{iPINNs on 1-D convection equation.}
    \label{fig:error_convection}
\end{figure}

In Figures \ref{fig:error_reaction} and \ref{fig:error_convection}, we illustrate the error of iPINNs on convection and reaction equations and the exact solutions for every value of parameter $\beta$ or $\rho$ that were considered.  Overall, we see that the neural network learns more complicated tasks more accurately if parts of the network are pretrained with easier tasks. At the same time, iPINNs replay the training data for previous PDEs during training for the new one. There are no additional costs to store or generate input points $(x, t)$ for previous tasks since they can be easily sampled when necessary.

Another illustration of the method is learning the problem \ref{eq:P2}. We consider the values of $\rho$ and $\nu$ for which PINN does not have difficulties with learning each component separately. Results obtained when first learning the reaction part (diffusion part) are shown in Table \ref{tab:rd} (Table \ref{tab:dr}). The main finding is that the network can learn almost every equation at least as well as when it is learned independently. In fact, for the reaction equation, the neural network improves significantly the prediction error. Another interesting observation is that the model learns the reaction-diffusion equation with almost the same error, regardless of the order of the tasks.  

\begin{table}[ht!]
    \caption{Final error and forgetting for reaction $\to$ diffusion $\to$ reaction-diffusion.}
        \centering
            \begin{tabular}{lcccc}
                \toprule
                 parameters & equation & & regular PINN & iPINN\\
                \midrule
                \multirow{6}{*}{$\rho = 3, \ \nu=5$} & \multirow{2}{*}{reaction} & abs. err & $6.72 \times 10^{-3}$ & $\mathbf{9.41 \times 10^{-4}}$\\
                                                                                 && rel. err & 2.05\% & $\mathbf{0.31\%}$\\
                                                    & \multirow{2}{*}{diffusion} & abs. err & $\mathbf{1.38 \times 10^{-4}}$ & $1.85 \times 10^{-4}$\\
                                                                                 && rel. err & $\mathbf{0.05\%}$ & 0.06\% \\   
                                                    & \multirow{2}{*}{reaction-diffusion} & abs. err & $4.89 \times 10^{-3}$ & $\mathbf{4.10 \times 10^{-3}}$\\
                                                                                 && rel. err & 0.80\% & $\mathbf{0.68\%}$ \\
                \midrule
                \multirow{6}{*}{$\rho = 4, \ \nu=4$} & \multirow{2}{*}{reaction} & abs. err & $1.13 \times 10^{-2}$ & $\mathbf{7.88 \times 10^{-3}}$\\
                                                                                 && rel. err & 3.68\% & $\mathbf{2.99\%}$ \\
                                                    & \multirow{2}{*}{diffusion} & abs. err & $\mathbf{4.35 \times 10^{-4}}$ & $5.84 \times 10^{-4}$\\
                                                                                 && rel. err & $\mathbf{0.16\%}$ & 0.19\%  \\   
                                                    & \multirow{2}{*}{reaction-diffusion} & abs. err & $4.58 \times 10^{-3}$ & $\mathbf{4.42 \times 10^{-3}}$\\
                                                                                 && rel. err & 0.70\% & $\mathbf{0.67\%}$\\ 
                \midrule
                \multirow{6}{*}{$\rho = 4, \ \nu=5$} & \multirow{2}{*}{reaction} & abs. err & $5.04 \times 10^{-2}$ & $\mathbf{4.20 \times 10^{-3}}$\\
                                                                                 && rel. err & 12.19\% & $\mathbf{1.71\%}$\\
                                                    & \multirow{2}{*}{diffusion} & abs. err & $5.18 \times 10^{-4}$ & $\mathbf{2.30 \times 10^{-4}}$\\
                                                                                 && rel. err & 0.18\% & $\mathbf{0.08\%}$\\   
                                                    & \multirow{2}{*}{reaction-diffusion} & abs. err & $4.61 \times 10^{-3}$ & $\mathbf{4.58 \times 10^{-3}}$\\
                                                                                 && rel. err & 0.69\% & $\mathbf{0.68\%}$\\ 
                                                                                 
            \bottomrule
            \end{tabular}
            \label{tab:rd}
\end{table}

\begin{table}[ht!]
    \caption{Final error and forgetting for diffusion $\to$ reaction $\to$ reaction-diffusion.}
        \centering
            \begin{tabular}{lcccc}
                \toprule
                 parameters & equation & & regular PINN & iPINN\\
                \midrule
                \multirow{6}{*}{$\rho = 3, \ \nu=5$} & \multirow{2}{*}{diffusion} & abs. err & $\mathbf{1.38 \times 10^{-4}}$ & $8.64 \times 10^{-4}$ \\
                                                                                 && rel. err & $\mathbf{0.05\%}$ & 0.28\%\\
                                                    & \multirow{2}{*}{reaction} & abs. err & $6.72 \times 10^{-3}$ & $\mathbf{2.11 \times 10^{-3}}$\\
                                                                                 && rel. err & 2.05\% & $\mathbf{0.68\%}$\\   
                                                    & \multirow{2}{*}{reaction-diffusion} & abs. err & $4.89 \times 10^{-3}$ & $\mathbf{4.07 \times 10^{-3}}$\\
                                                                                 && rel. err & 0.80\% & $\mathbf{0.67\%}$\\
                \midrule
                \multirow{6}{*}{$\rho = 4, \ \nu=4$} & \multirow{2}{*}{diffusion} & abs. err & $4.35 \times 10^{-4}$ & $\mathbf{3.45 \times 10^{-4}}$\\
                                                                                 && rel. err & 0.16\% & $\mathbf{0.12\%}$\\
                                                    & \multirow{2}{*}{reaction} & abs. err & $1.13 \times 10^{-2}$ & $\mathbf{4.91 \times 10^{-3}}$\\
                                                                                 && rel. err &  3.68\% & $\mathbf{1.97\%}$\\   
                                                    & \multirow{2}{*}{reaction-diffusion} & abs. err & $4.58 \times 10^{-3}$ & $\mathbf{4.42 \times 10^{-3}}$\\
                                                                                 && rel. err & 0.70\% & $\mathbf{0.67\%}$\\ 
                \midrule
                \multirow{6}{*}{$\rho = 4, \ \nu=5$} & \multirow{2}{*}{diffusion} & abs. err & $\mathbf{5.18 \times 10^{-4}}$ & $1.05 \times 10^{-3}$\\
                                                                                 && rel. err & $\mathbf{0.18\%}$ & 0.33\%\\
                                                    & \multirow{2}{*}{reaction} & abs. err &   $5.04 \times 10^{-2}$ & $\mathbf{9.15 \times 10^{-3}}$ \\
                                                                                 && rel. err & 12.19\% & $\mathbf{3.39\%}$\\   
                                                    & \multirow{2}{*}{reaction-diffusion} & abs. err & $4.61 \times 10^{-3}$ & $\mathbf{4.30 \times 10^{-3}}$\\
                                                                                 && rel. err & 0.69\% & $\mathbf{0.65\%}$\\ 
                                                                                 
            \bottomrule
            \end{tabular}
            \label{tab:dr}
\end{table}

\section{Additional study}

In this section, we provide additional information about the learning procedure of iPINNs . We highlight some important training details such as the presence of regularization and the choice of activation functions. Also, we explore the subnetworks that our approach produces showing the proportion of parameters allocated to each task.

\subsection{Sensitivity to hyperparameters} \label{sec:hyperparameters}

\begin{figure}[ht!]
    \centering
    \includegraphics[width=\textwidth]{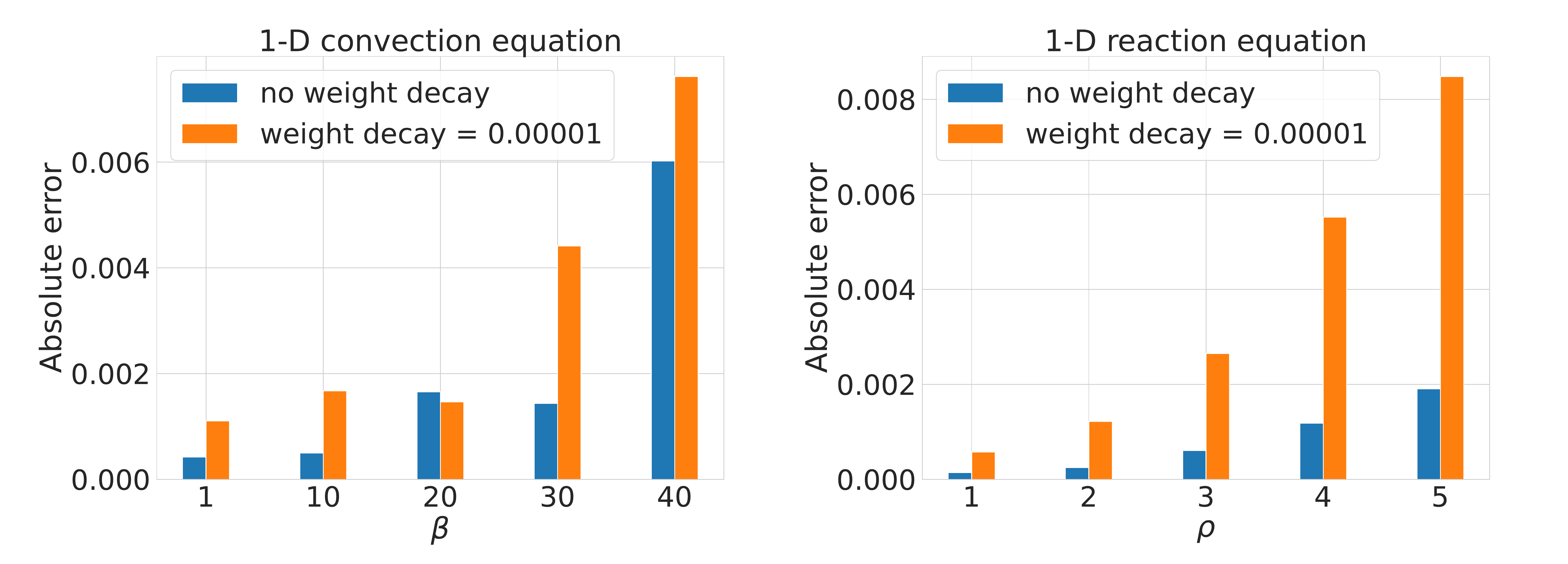}
    \caption{Influence of weight decay on the results for reaction (left) and convection (right) equations after all tasks are learned.}
    \label{fig:wd_comparison}
\end{figure}

Here we illustrate the influence of different training hyperparameters on the performance of iPINNs. First, we compare the results with and without regularization parameter (weight decay). In Figure \ref{fig:wd_comparison}, it can be observed that the presence of weight decay worsens the prediction error. However, looking at the result it is clear that iPINNs still work if weight decay is present. We can explain the lack of need for weight decay with the fact that many parameters are assigned to multiple tasks and cannot overfit to a particular one. Each subnetwork is also less parameterized than the original network and therefore does not tend to overfit. Thus, weight decay is not necessary and its presence only worsens the result due to the complication of the optimization procedure.

\begin{figure}[ht!]
    \centering
    \includegraphics[width=\textwidth]{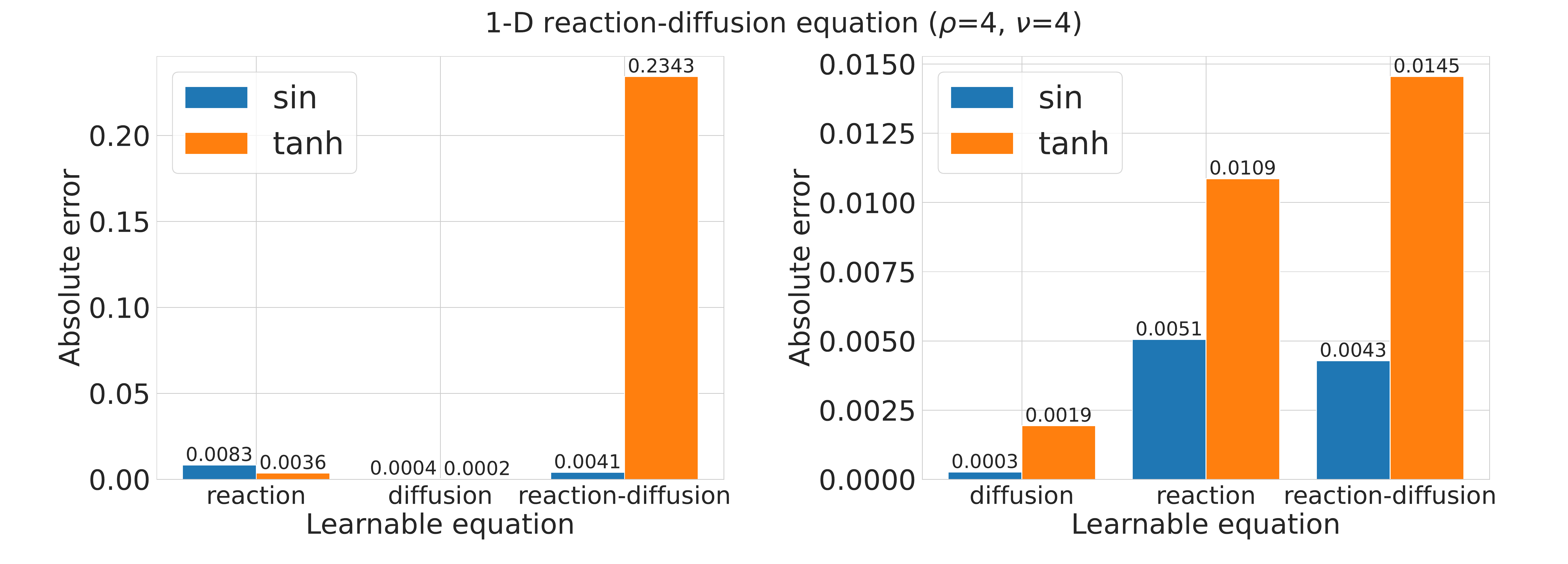}
    \caption{Influence of activation function on the results when the reaction learned first (left) and diffusion learned first (right).}
    \label{fig:activation_comparison}
\end{figure}

Furthermore, we compare the performance when using \texttt{sin} and \texttt{tanh} activation functions for two task orderings in Figure \ref{fig:activation_comparison}. We observe that \texttt{sin} works significantly better in both cases. Also, we test \texttt{ReLU} activation but it demonstrates poor performance in both PDEs orderings. If the reaction is learned first, the absolute errors are $0.4959, 0.2369$ and $0.1493$. If we start with the diffusion equation and then learn reaction and reaction-diffusion PDEs, the errors are $0.2399, 0.2977$ and $0.3003$.

\begin{figure}[ht!]
    \centering
    \includegraphics[width=\textwidth]{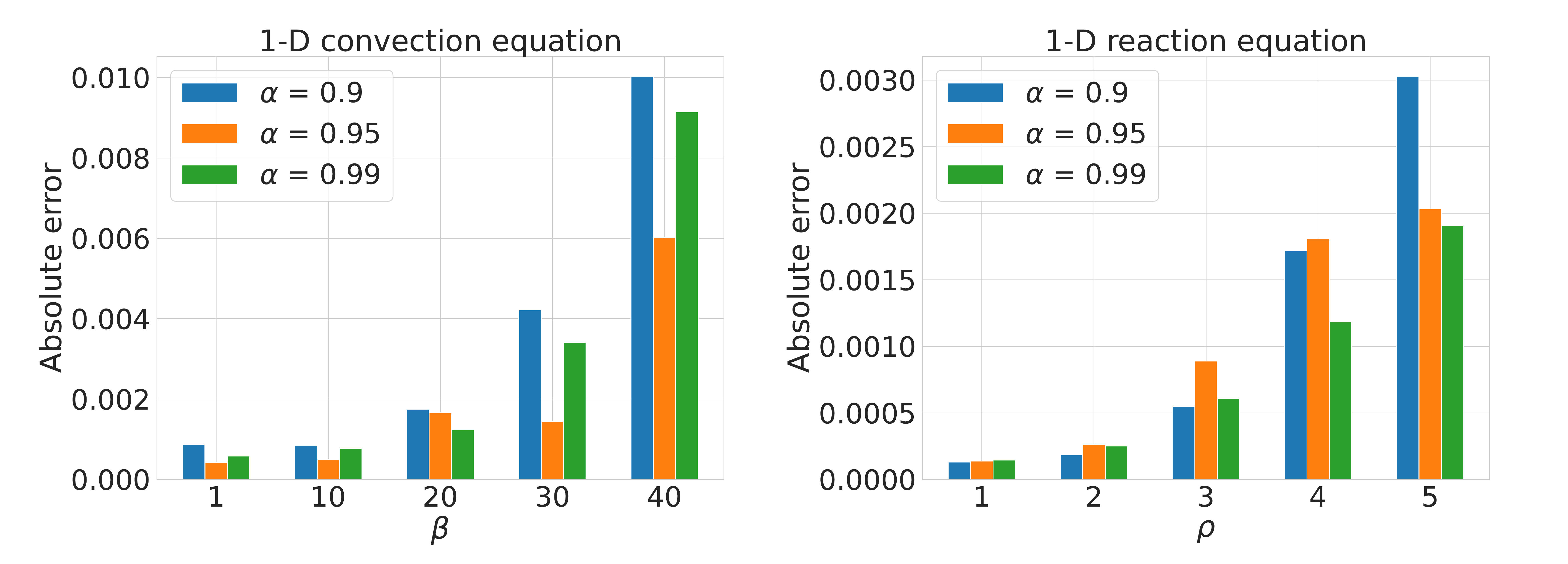}
    \caption{iPINNs with different values of pruning parameter $\alpha$.}
    \label{fig:alpha_comparison}
\end{figure}

In addition, we present how different values of pruning parameter $\alpha$ affect the results. The higher the value of $\alpha$ is, the less the network is pruned. Therefore, if $\alpha = 0.95$ the  task-specific subnetworks are sparser than with $\alpha=0.99$ but less sparse if $\alpha = 0.9$. In Figure \ref{fig:alpha_comparison}, we observe that for the reaction equation, we can prune less and achieve better performance which can be explained by the fact that PDEs in the reaction family are quite similar. Therefore, we can allow the network to have more overlaps to share knowledge between subnetworks. For the case of learning within the same family of convection PDEs, the value of $\alpha = 0.95$ was revealed to be a better option for constructing a sufficiently expressive task-specific subnetwork and frees space for future tasks. Notwithstanding, the performance is good with any reasonable choice of pruning parameter.

\subsection{Subnetworks analysis}

\begin{figure}[ht!]
    \begin{subfigure}{0.45\textwidth}
        \centering
        \includegraphics[width=\textwidth]{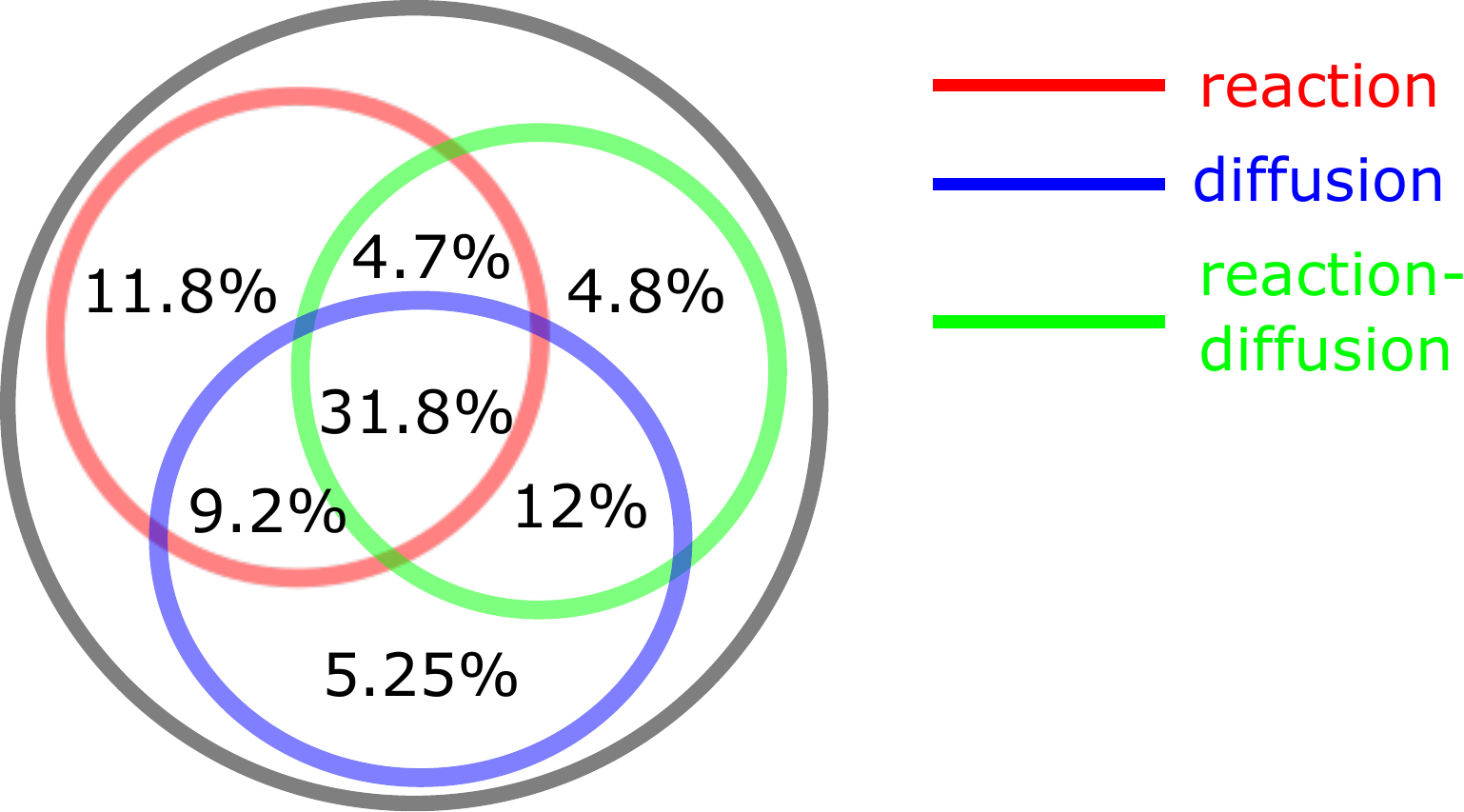}
        \caption{reaction $\to$ diffusion $\to$ reaction-diffusion.}
        \label{fig:r-d-rd44_masks}
    \end{subfigure}\hfill
    \begin{subfigure}{0.45\textwidth}
        \centering
        \includegraphics[width=\textwidth]{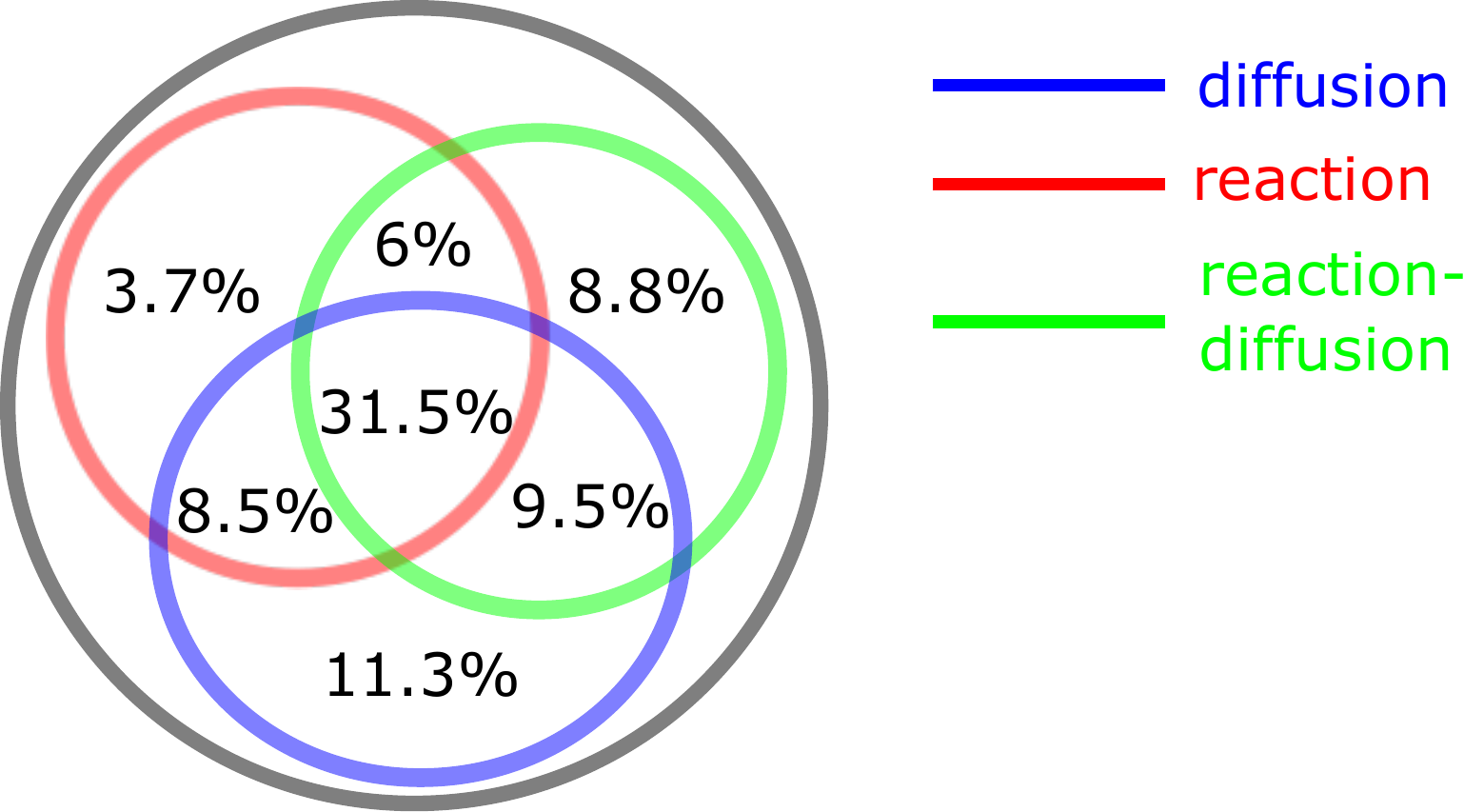}
        \caption{ diffusion $\to$ reaction $\to$ reaction-diffusion.}
        \label{fig:d-r-dr44_masks}
    \end{subfigure}
    \caption{Percentage of parameters used for every equation with $\rho=4, \ \nu=4$.}
    \label{fig:rd44_masks}
\end{figure}

In Figure \ref{fig:rd44_masks}, we present the portions of the subnetworks that are occupied by each task. We will illustrate this by considering both orders -- when the model learns the reaction equation first (Figure \ref{fig:r-d-rd44_masks}), and when diffusion comes first (Figure \ref{fig:d-r-dr44_masks}). These results are averaged over 3 different runs for each of the orderings. It is noteworthy that the percentage of parameters occupied by all tasks is very similar for both orderings (31.8\% and 31.5\% respectively of all network parameters). On the other hand, the percentages of used parameters for both cases are 79.5\% and 79.3\%. This means that the total number of trained parameters for the two incremental procedures is the same for both cases, which shows the robustness of the method. Moreover, the network has about 20\% of free connections to learn new tasks. 

\section{Conclusion}

In this work, we propose an incremental learning approach for PINNs where every task is presented as a new PDE. Our algorithm is based on task-related subnetworks for every task obtained by iterative pruning. To illustrate our idea, we consider two cases when incremental learning is applicable to a sequence of PDEs. In the first case, we consider the family of convection/reaction PDEs, learning them sequentially. In the second example, we consider the reaction-diffusion equation and learn firstly the components of the process, namely reaction and diffusion, and only then the reaction-diffusion equation. Our main goal is to show the possibility of incremental learning for PINNs without significantly forgetting previous tasks. From our numerical experiments, the proposed algorithm can learn all the given tasks, which is not possible with standard PINNs. Importantly, we also show that future tasks are learned better because they can share connections trained from previous tasks, leading to significantly better performance than if these tasks were learned independently. We demonstrate that this stems from the transfer of knowledge occurring between subnetworks that are associated with each task. Interestingly, the model's performance on previous tasks is also improved by learning the following tasks. In essence, iPINNs demonstrate symbiotic training effects between past and future tasks by learning them with a single network composed of dedicated subnetworks for each task that share relevant neuronal connections.

\bibliographystyle{splncs04}
\bibliography{references}

\end{document}